\documentclass{article}

\usepackage[main,final]{neurips_2026}

\usepackage[utf8]{inputenc} %
\usepackage[T1]{fontenc}    %
\usepackage{hyperref}       %
\usepackage{url}            %
\usepackage{booktabs}       %
\usepackage{amsfonts}       %
\usepackage{nicefrac}       %
\usepackage{microtype}      %
\usepackage{xcolor}         %

\usepackage{amsmath}
\usepackage{amssymb}
\usepackage{mathtools}
\usepackage{amsthm}
\usepackage{paralist}       %
\usepackage[capitalize,noabbrev]{cleveref}  %
\usepackage[most]{tcolorbox}

\usepackage[textsize=tiny]{todonotes}
\newcommand{\note}[1]{\todo[inline]{#1} }
\renewcommand{\note}[1]{}  %

\newcommand{\ours}{\texttt{SZP}}

\newcommand{\mup}{\mu\text{P}}
\newcommand{\muP}{\mu\text{Parameterization}}

\newcommand{\nton}{\texttt{Net2Net}}
\newcommand{\hc}{\texttt{HyperCloning}}
\newcommand{\btob}{\texttt{bert2BERT}}
\newcommand{\WS}{\texttt{Warmstarting}}
\newcommand{\ws}{\texttt{WS}}
\newcommand{\FP}{\textit{function-preservation}}
\newcommand{\fp}{\texttt{FP}}
\newcommand{\scratch}{\texttt{Scratch}}
\newcommand{\tkpm}{\textit{tokens/parameter}}

\newcommand{\snp}{{\textit{shrink-and-perturb}}}
\newcommand{\SNP}{\textit{SnP}}

\newcommand{\growth}{g}
\newcommand{\gub}{g_\text{upper}}
\newcommand{\gopt}{g_\text{opt}}
\newcommand{\Growth}{\mathcal{G}}

\newcommand{\shrinkhp}{\lambda_{\text{shrink}}}
\newcommand{\perturbhp}{\sigma_{\text{perturb}}}

\title{When is Warmstarting Effective \\for Scaling Language Models?}

\author{%
\begin{tabular}{c}
Neeratyoy Mallik$^{1,2,*}$ \quad
Maciej Janowski$^{1,3,*}$ \quad
Johannes Hog$^{1,*}$ \\
Herilalaina Rakotoarison$^{4}$ \quad
Josif Grabocka$^{3}$ \quad
Frank Hutter$^{5,6,1}$ \quad
Aaron Klein$^{6}$ \\[0.6em]
\small $^{1}$University of Freiburg \quad
\small $^{2}$Zuse School ELIZA \quad
\small $^{3}$University of Technology Nuremberg  \\
\small $^{4}$University of Helsinki \quad
\small $^{5}$Prior Labs \quad
\small $^{6}$ELLIS Institute T\"ubingen \quad \\
\small $^{*}$Equal contribution. \\
\small Correspondence: \texttt{\{mallik,janowski,hogj\}@cs.uni-freiburg.de}
\end{tabular}
}

\begin{document}

\maketitle

\begin{abstract}
Model growth from a given checkpoint aims to accelerate training of a larger model, offering potential resource savings.
Despite recent interest, warmstarting has seen limited practical adoption in large-scale training.
We attribute this to two underexplored factors: (1) an overemphasis on preserving the smaller model's performance at initialization, which constrains operator design for new architectures, and (2) insufficient analysis of how growth interacts with hyperparameters and scaling behavior, compounded by inconsistent growth factors across the literature.

We show that preserving the base model's initial post-growth performance is not necessary for strong final performance, and that simple, architecture-agnostic growth strategies can outperform more complex warmstarting operators.
Crucially, we empirically identify an upper bound on the growth factor $g$ beyond which training from scratch is more efficient.
We observe this across multiple ablation setups. 
Notably, this limit is also present, but unreported, in prior published results.
Across our experiments on dense MLPs and dense language models, we find that a $2\times$ growth factor is the most reliable in yielding convergence speedups, with gains most pronounced under 20 \tkpm{} budgets and diminishing as budget increases.
We fit scaling laws over these observations to provide guidance for practitioners deciding when and how much to grow.
Together, our analysis provides practical guidelines and empirical limits for model growth.
\end{abstract}

\section{Introduction}\label{sec:introduction}

Training state-of-the-art deep learning models requires significant computational resources. 
As models scale from millions to billions of parameters, practitioners face an important question: can we reuse the compute invested in training smaller models when scaling up? 
Model growth or warmstarting (\ws{}), i.e., initializing larger models from trained smaller checkpoints, offers potential efficiency gains, yet remains largely absent from production pipelines despite substantial research~\citep{gong-icml19a,chen-acl21a,rae-arxvi21a,shen-pmlr2022,wang-iclr23a,wang-icml23a,deng-iccv23a,samragh-arxiv24a,du-neurips24a,yu-arxiv26a}.

Deep Learning (DL) success has been built on pushing the loss frontier across tasks, applications, and data modalities~\citep{krizhevsky-nips12a,he-cvpr16a,vaswani-neurips17a,brown-neurips20a,radford-icml21a,rombach-cvpr22a,openai-24}. 
As training efficacy improves, processing larger datasets drives corresponding increases in model capacity~\citep{alabdulmohsin-neurips23a}. 
Current research on frontier models builds on seminal scaling law studies demonstrating the necessity of simultaneously growing both model size and data processed~\citep{kaplan-arxiv20a,hoffmann-neurips22a}.
This co-scaling question continues to evolve across different constraints, modalities, and architectures, with the traditional goal of pushing the loss-frontier now complemented by optimizing the compute-optimal frontier~\citep{anagnostidis-icml24a}. 
In practice, these considerations translate to decisions about resource efficiency~\citep{wan2023efficient}.

While \ws{} has been periodically explored across architectures and paradigms, it has not become standard practice in training pipelines, often remaining an academic topic rather than a common choice in DL pipelines~\citep{gong-icml19a,wang-iclr23a}. 
We identify three barriers to practical \ws{} adoption: 
\begin{inparaenum}[1)]
    \item Implementation complexity: existing methods require architecture-specific adaptations and may not generalize to modern components~\citep{su-arxiv21a,gu-arxiv23a};
    \item Unclear growth limits: practitioners lack guidance on how much a given checkpoint can be grown before \ws{} becomes counterproductive~\citep{xie-eccv22a,wang-iclr23a};
    \item Hyperparameter opacity under scaling: the interaction between \ws{} methods, hyperparameters, and model and data scaling budgets is under-explored, creating tuning overhead that undermines efficiency gains~\citep{fetterman-23a,dey-arxiv25a}.
\end{inparaenum}
Moreover, in scaling regimes where both models and data grow to improve final convergence loss, measuring relative early-stage convergence speed may not adequately capture practical requirements~\citep{shin-icml24a,unlu-arxiv26a}.

In this work, we attempt to systematically investigate when \ws{} is truly effective, in a first-principles manner, to address these barriers.
We limit the discussion scope, both for clarity and computation, for the case when a \textit{base} model checkpoint is available to be \ws-ed to a larger model scale. 
We assume that we are aware of the dataset the \textit{base} was trained on, and that some of its training hyperparameters may be available, but we treat it as a \textit{black-box} when it comes to the training compute invested in it (unlike the assumption made by~\citet{liew-arxiv25a}). 
We believe this better reflects a real-world scenario where only model checkpoints are publicly released.
Our contributions are:

\begin{enumerate}[1.]
    \item We generalize warmstarting growth into \emph{shrink-zero-perturb} (\ours{}), a simple architecture-agnostic operation, and find that exact function-preservation is not required for effective warmstarting.

    \item We empirically establish an upper bound on the growth ratio of a given checkpoint, such that, beyond this bound, training the grown model from scratch converges as fast as warmstarting it.

    \item Using scaling-law fits, we show that for a given warmstarting method and base model size, the convergence advantage over training from scratch persists only within a bounded token budget, and suggest the fitted crossover as a practical guide for deciding when to warmstart.

\end{enumerate}

The paper proceeds as follows: \Cref{sec:related_bg} reviews prior warmstarting methods and guidance; \Cref{sec:methodology} introduces \ours{} and its design motivation; \Cref{sec:empirical} reports hyperparameter transfer, growth limits, and scaling-law predictions; and \Cref{sec:conclusion} summarizes practical guidelines and limitations. 

We release our training pipelines in the linked repository\footnote{\url{https://github.com/Neeratyoy/warmstarting_exps}
}.

\section{Related Works}\label{sec:related_bg}

\textbf{Model growth as initialization} 
is the broad class of \ws{} techniques that derive the weights of a larger model from a smaller, typically well-trained checkpoint, before beginning training.
Here, the aim is to transfer the knowledge learned by the smaller model to improve the larger model's training efficiency.
\nton{}~\citep{chen-iclr16a} and Network Morphisms~\citep{wei-icml16a} are among the earliest works exploring model growth for Deep Learning. 
At the time, the scope was limited to convolutional networks, but the core idea was carried forward to evolving architectures~\citep{gong-icml19a,chen-acl21a,yao-iclr24a,samragh-arxiv24a}.
The benefit lies in being a one-time operation before training, requiring only weights from a smaller network of the same architectural family.
A central idea in designing \ws{} operations for initialization is that of \FP{} (\fp{}), wherein the grown model's output at the first forward pass matches that of the base model checkpoint used for \ws{}.
Introduced by \citet{chen-iclr16a} for Deep Learning, \fp{} has dominated the width-scaling \ws{} paradigm~\citep{wang-icml23a,samragh-arxiv24a,du-neurips24a,ma-arxiv26a}.
However, with new architectures, it typically requires special adaptation to remain function preserving~\citep{chen-acl21a,samragh-arxiv24a}.
We thus question if \fp{} at initialization is a requirement for model growth.
~\citet{karp-arxiv24a} already showed that loss after a short training budget correlates better than loss at initialization with growth operator rankings.
We extend this to more practical training budgets.

\textbf{Learned model growth} is the other family of \ws{} that does not fix the exact manner of \ws{} operation but handles it dynamically.
~\citet{wang-iclr23a} introduces a Kronecker-factored linear operator that parameterizes the linear map from the small to the larger model, learned over a few steps of training (in the style of ~\citet{karp-arxiv24a}).
Here, the Kronecker product constrains growth ratios but works for both width and depth.
~\citet{evci-iclr22a} looks at dynamic model growing during training by maximizing the gradient norms of the newly added neurons to enforce faster convergence, while the output-preservation in the \FP{}-style is still implied.
The learned component here comes from the newly grown weights set through SVD over the gradient on the activations.
This design looks to prevent dead neurons in grown models, which can lead to early rank collapse or not utilize the grown parameter capacity and move sufficiently away in the function hypothesis space from the base model.
This connects to plasticity loss~\citep{lyle-icml23a}, which correlates with dead (inactive) neurons, low effective rank representations, and growing weight magnitudes.
~\citet{dohare-arxiv23a} addresses this for a trained model by selectively reinitializing a fraction of low-utility neurons, scored by their activations, after each training step to keep a model in a continual learning setup.
In this work, we focus entirely on the more straightforward setup of initialization-only \ws{}.

\textbf{Full-state \WS{}} looks beyond the weights.
~\citet{wang-iclr2024a} looks at the learning rate schedulers,~\citet{ma-arxiv26a} and~\citet{yu-arxiv26a} necessitate the \ws{} of optimizer states too.
These families aim to preserve loss (and thereby \fp{}), while suitably inducing training dynamics through optimizer schedules and states, including hyperparameters.
Our setup assumes that only the smaller model checkpoint is available and no other information on the optimizer or scheduler, especially their states.
The framework of investigation we offer in our paper should apply to future studies extending to complete training state warmstarts, too.

\textbf{\WS{} in practical settings} has been covered mostly in terms of operator selection~\citep{karp-arxiv24a}, for scaling related recommendations involving scaling laws~\citep{wang-icml23a,liew-arxiv25a}, and the role of width-depth in \ws{} efficacy~\citep{du-neurips24a}.
We note that most \ws{} literature varies in their setups, especially with respect to growth factors $g$ (the ratio of target to base model parameters) and the budget involved in training.
Our investigation in this work focuses specifically on width-scaling strategies.
Typically, joint scaling of models along both dimensions requires independent choice of width- and depth-growth operators~\citep{du-neurips24a}. 
Our study here is compatible with any depth-growing and thereby joint-growing setups.
We adopt a ``black-box'' assumption where the tokens spent on the base checkpoint are unknown, and we are concerned primarily with the question of how much we can grow this model, by training for how much budget, to ensure there is a benefit over training from scratch.
In the remaining paper, we treat the terms \textit{Model Growth} and \WS{} (\ws{}) synonymously.

Refer to Appendix~\ref{app:related} for additional literature on hyperparameter transfer and scaling laws.

\section{Methodology}\label{sec:methodology}

This section defines the \ws{} (\WS{}) setup for our empirical study (\Cref{sec:empirical}). 
We first introduce a compact design space for growth (\Cref{sec:design-space}) that abstracts common warmstarting operators. 
We then use this space to identify a robust, architecture-agnostic default in~\ours{} (\Cref{sec:szp-choice}).

\subsection{Design Space for Warmstarting}\label{sec:design-space}

To study when \WS{} is effective, we compare simple, architecture-agnostic \ws{} operations against training from scratch.
Function-preservation (\fp{}) methods typically combine three ingredients: cloning existing weights, scaling grown weights, and adding noise, often through architecture-specific rules~\citep{chen-iclr16a,chen-acl21a,du-neurips24a,samragh-arxiv24a,ma-arxiv26a} (see Appendix~\ref{app:net2net}). 
We instead keep these interacting design choices independent and apply them uniformly across all components, both $1$D and $2$D weights.

We define our design space by extending the \snp{} (\SNP{}) formulation from Continual Learning (CL)~\citep{ash-neurips20}. 
In its original form, target weights $\theta_{\text{target}}$ are derived by rescaling the base weights and adding Gaussian noise:

\begin{equation}
    \theta_{\text{target}}^{l}~=~\shrinkhp{} \cdot \theta_{\text{base}}^{l} + \mathcal{N}(0, \perturbhp{}^2)
    \label{eq:vanilla-snp}
\end{equation}
where $\shrinkhp{}$ regulates the influence of inherited weights and $\perturbhp{}$ controls the standard deviation of Gaussian noise, used as perturbation for all neurons.

To adapt this for model growth, where the target model is larger than the base, we augment (in \underline{\color{blue}blue}) the formulation with a growth function, $\Growth$:
\begin{equation}
    \theta_{\text{target}}^{l}~=~\shrinkhp{} \cdot \underline{\color{blue}\Growth(\theta_{\text{base}}^{l}, p, q)} + \mathcal{N}(0,\perturbhp{}^2),
    \label{eq:growth-snp}
\end{equation}
where $\Growth{}: \mathbb{R}^{m \times n} \times \mathbb{N} \times \mathbb{N} \rightarrow{} \mathbb{R}^{p \times q}$ maps the base weights $\theta_{\text{base}}^{l}~\in~\mathbb{R}^{m \times n}$ into the larger target shape $p \times q$ ($p \ge m, q \ge n$).
Here, $\shrinkhp{} \in \mathcal{R} > 0$ and $\perturbhp{} \in \mathcal{R}$, and $l\in\mathcal{N}$ denoting layer index.

\paragraph{Intuition and Motivation.}

By separating $\Growth$, $\shrinkhp{}$, and $\perturbhp{}$ in a \ws{} operations, it gives a lightweight abstraction of common \fp{}-style operations without requiring component-specific rules for exact function preservation.
Formally,~\Cref{eq:growth-snp} can represent more complex operators like \nton{} through appropriate choices of $\Growth$, $\shrinkhp{}$, and $\perturbhp{}$, per architecture component~\citep{samragh-arxiv24a}, though deriving such mappings remains intentionally out of scope here.

Conceptually, \textit{Growth} determines which structure is inherited from the base model. \textit{Perturbation} is critical for symmetry breaking, that is, it ensures that newly added neurons can learn distinct features rather than simply duplicating inherited behavior. 
Finally, \textit{Shrinking} reduces the magnitude of inherited weights, dampening the dominance of base representations, and allows the new neurons to participate meaningfully in learning.
Together, these axes yield simple \ws{} realizations that connect cleanly to established CL concepts while remaining easy to implement. 
We offer theoretical intuition in Appendix~\ref{app:szp-theory} and perform a mechanistic analysis of this setup in Appendix~\ref{app:interpretability}.

\subsection{Choosing a Default Warmstarting Method}\label{sec:szp-choice}

\begin{figure}[htbp]
    \centering
    \includegraphics[width=0.95\linewidth]{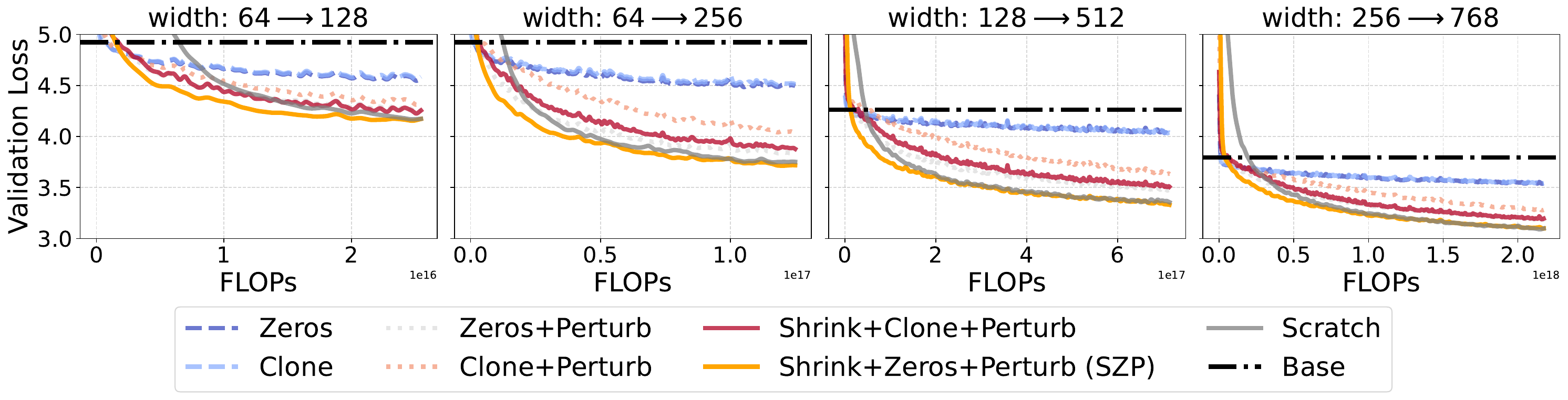}
    \caption{Comparing the simplest set of realizations of~\Cref{eq:growth-snp}, for different \textit{target} scale runs, given the \textit{base} checkpoint (\textit{base} $\rightarrow$ \textit{target});
    \texttt{Zeros} and \texttt{Clone} represent the choices for $\Growth$, the growth function;
    \texttt{Zeros}$+$/\texttt{Clone}$+$\texttt{Perturb} represents the addition of noise to the grown matrix;
    Adding \texttt{Shrink} to the inherited base model implies a scalar multiplication of $\shrinkhp{}$ to the \textit{base} matrix before any growth operation is applied.
    For either choice of $\Growth$, \texttt{Shrink}$+ \Growth +$\texttt{Perturb} converges faster than $\Growth +$\texttt{Perturb}, which converges faster than $\Growth$ alone.
    \texttt{Scratch} runs the target model with standard initialisation (no warmstarting), while \texttt{Base} marks the final loss of the smaller base model used for \ws{}.
    All \ws{} models appear to recover and go past the \texttt{Base} loss quickly; only \ours{} keeps a convergence rate better or matched with the vanilla run from \texttt{Scratch}.
    }
    \label{fig:basic-baselines}
\end{figure}

We compare representative realizations of the design space in~\Cref{fig:basic-baselines} to select a robust default for our scaling experiments. 
In~\Cref{fig:basic-baselines}, \texttt{Zeros} and \texttt{Clone} show no significant difference in isolation, but each improves with perturbation noise (\texttt{Zeros+Perturb}, \texttt{Clone+Perturb}), and further with shrinking (\texttt{Shrink+Zeros+Perturb}, \texttt{Shrink+Clone+Perturb}) across growth ratios.
We view these variants as streamlined realizations of \ws{} that remain architecture-agnostic and easy to implement.
A function-preservation analogue would instead set the scaling magnitude relative to grown sizes, apply a weight-type-specific cloning operation, and similarly adjust the perturbation magnitude (e.g. \nton{}~\citep{chen-iclr16a}, \hc{}~\citep{samragh-arxiv24a}).

Notably, all \ws{} variants recover the base-model loss within a few steps and converge faster than \scratch{} early in training.
However, the differences between variants become clearer in sustained convergence, where \ours{} (\texttt{Shrink+Zeros+Perturb}) is the only variant consistently superior across model widths. 
We therefore use \ours{} as our main architecture-agnostic \ws{} method.

While simplicity is subjective, \ours{} strikes a practical middle ground: it mirrors the structure of standard \fp{} algorithms while loosening the strict constraint of preserving base-model outputs at initialization.
We emphasize that \ours{} is not proposed as a new state-of-the-art warmstarting algorithm, but as a principled baseline we find effective for studying the scaling behavior of \ws{} in~\Cref{sec:empirical}.

\section{Empirical Study}\label{sec:empirical}

In this section, we use the \ours{} setup to study its scaling behavior, compare against popular \FP{} baselines such as \nton{}, observe the existence of an upper bound on the growth factor, and fit scaling laws to characterize when training from scratch outperforms \WS{} (\ws{}). 
We distill these findings into practical recommendations.

\subsection{Experiment Setup}\label{sec:empirical-setup}

Following~\citet{qiu-icml25a}, we use the synthetic regression task as a controlled testbed for scaling and \ws{} (refer to Appendix~\ref{app:synthetic-regression} for more details).
This setting is computationally light to afford dense hyperparameter sweeps at both \emph{base} and \emph{target} widths, for all run types: \scratch{}, and \ws{} runs (\nton{} and \ours{}).
The language modeling setup, used for the broader empirical comparison, is described below. We report the corresponding compute estimates in Appendix~\ref{app:compute}.

\subsubsection{Hyperparameter Selection}\label{sec:exp-hps}
When treating \ws{} as an initialization, it is expected to reshape the loss landscape over hyperparameters.
Prior comparisons of \ws{} operations have under-explored this.
This matters particularly when \ws{} to much larger models, where re-tuning costs can be prohibitive.
Here, we use the synthetic MLP regression task to compare tuning strategies across \ws{} operations.

\begin{figure}[hbtp]
    \centering
    \includegraphics[width=0.95\linewidth]{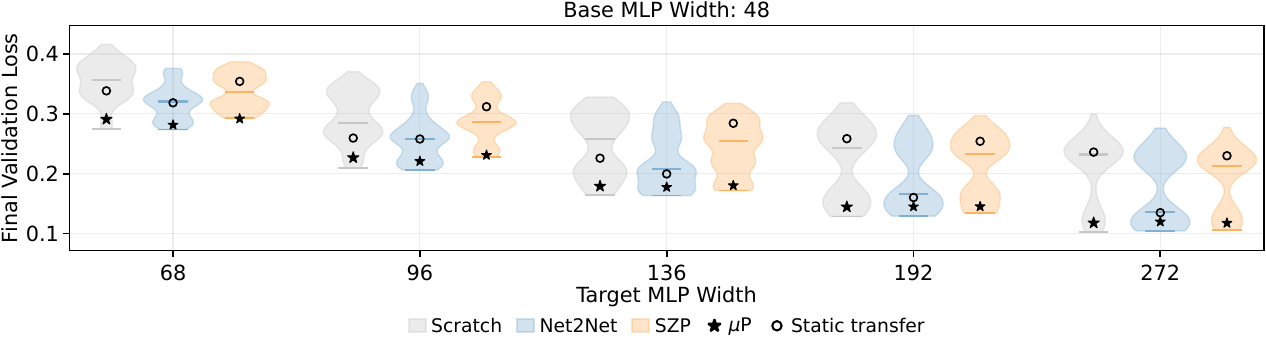}
    \caption{
    Final validation loss under MLP-width scaling from a fixed base width of $48$.
    For each \emph{target} width, each triplet of violins shows Scratch, \nton{}, and \ours{} from left to right.
    Violins summarize the final-loss distribution over a dense target-scale hyperparameter grid (Appendix, \cref{tab:hyperparameter-search-space}); horizontal ticks denote the median and minimum losses within each grid.
    Open circles mark \textbf{static transfer}: the best base-width hyperparameters unchanged at the target width.
    Stars mark \textbf{width-aware transfer}: the base-width optimum follows $\mup{}$ scaling rules (Appendix~\ref{app:regression_hp_transfer}).
    }
    \label{fig:violin-hp}
\end{figure}

\paragraph{Training Hyperparameters.} The tuned hyperparameters are detailed in~\Cref{tab:hyperparameter-search-space} and summarized in~\Cref{fig:violin-hp}, where violins show the final validation loss distribution from a dense grid search at each target width for \scratch{}, \nton{}, and \ours{}.
For transfer, we take the best configuration at the base width of $48$ and transfer it as (1) unchanged (\emph{static transfer}), or (2) with width-aware $\mup$ scaling rules~\citep{yang-neurips21a}. 
We refer readers to Appendix~\ref{app:mup-details} for background and details on $\mup$.

\Cref{fig:violin-hp} shows that dense tuning at the target width usually yields the \textit{best} configuration when the budget allows. 
The effectiveness of transfer, however, depends on the initialization method. 
Static transfer worsens with growth factor for both \scratch{} and \ours{}, while \nton{}'s \FP{} aspect makes it more robust.
Crucially, $\mup$ transfer is consistently near-optimal relative to the dense grid and outperforms static transfer across all run types.
This motivates us to use $\mup{}$ transfer for all subsequent compute-heavy experiments, especially the language model runs reported in later sections.
Notably, the best attainable losses across dense grid search over \scratch{}, \nton{}, and \ours{} converge as the growth factor increases. 
This hints at the practical limits of \ws{} that we examine more closely in \Cref{sec:empirical-practice}.

\paragraph{\ours{} Design Choices.}
In \FP{}-based methods, the perturbation noise and scaling applied to base or new neurons depend on the architectural component.
We instead apply~\Cref{eq:growth-snp} uniformly across all weights and thus fix the scaling variable and the perturbation noise form.
For perturbation noise, we use the scale-dependent $\sigma_{\mathrm{perturb}}=1/\sqrt{\mathrm{width}}$ applied under both Static transfer and $\mu$P.
This choice is common in Deep Learning and has also been used in \fp{}-based setups~\citep{ma-arxiv26a}.
We provide an ablation on the perturbation noise in Appendix~\ref{app:perturbation-ablation}, where we observe that the noise scale affects the convergence rate more than the final generalization loss.

Just as perturbation is essential for new neurons to break symmetry inherited from the base model and use the full target capacity~\citep{yu-arxiv26a}, shrinking is essential for reducing weight norms, increasing effective ranks, and maintaining plasticity.
We follow~\citet{ash-neurips20} and fix $\shrinkhp{}=0.4$ in our experiments to isolate the effect of other variables.
Our ablation in Appendix~\ref{app:shrinking-ablation} confirms that $\shrinkhp{} \in \{0.2,~0.4\}$ offers improved performance over training from scratch for reasonable growth factors.
We additionally provide a simple theoretical sketch in Appendix~\ref{app:szp-theory} for why zero-padding, perturbation, and shrinkage play complementary roles.

\subsubsection{Language Model Setup}\label{sec:exp-lm-setup}
All language-model experiments train decoder-only GPT-2 style models~\citep{radford-openaiblog19a} using the open-source \textsc{LitGPT} implementation~\citep{litgpt23} on the \textsc{SlimPajama} corpus~\citep{slimpajama23} (\cref{fig:basic-baselines} uses sampled version\footnote{\url{https://huggingface.co/datasets/DKYoon/SlimPajama-6B}}).
We perform only width-scaling here, consistent with our scope of studying \fp{}-based \ws{}.
All runs used AdamW and a warmup-stable-decay learning rate scheduler with warmup and cooldown lengths set to fixed fractions of the total training budget.
Each run's training budget is set to $10, ~20,~\text{or}~30$ \tkpm{}.
We train models ranging from $14$M to $1.2$B parameters.
Except for dense grid searches at multiple base scales, all \scratch{} and \ws{} runs at scales larger than the base use $\mup{}$ transfer for learning rates, and we scale the base-scale optimal batch size.
Additional training and hyperparameter details are provided in Appendix~\ref{app:llm-details}.
We additionally present ablations of \ours{} performance on the LLaMa architecture in Appendix~\ref{app:llama-metrics}.

\subsection{Practical \WS{}}\label{sec:empirical-practice}

In this section, we focus on language model experiments and address our primary question of how and where to find gains when \WS{}.

\begin{figure}[htbp]
    \centering
    \begin{tabular}{c}
        \includegraphics[width=0.95\linewidth]{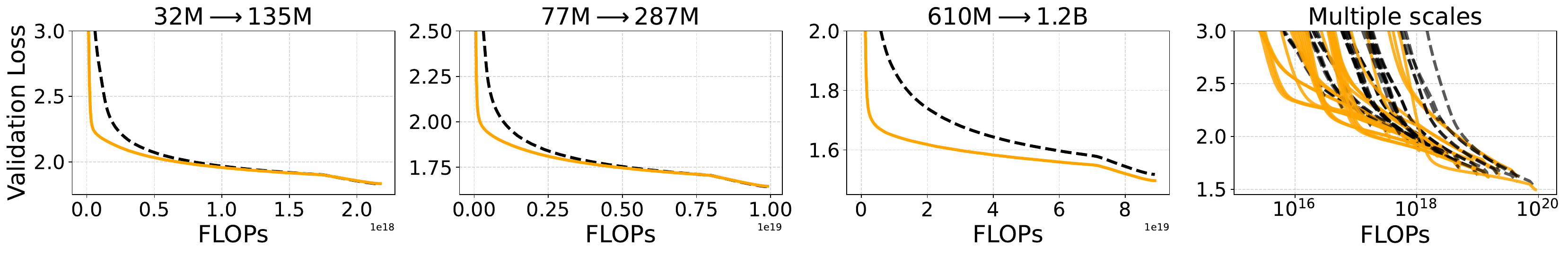} \\
        \includegraphics[width=0.95\linewidth]{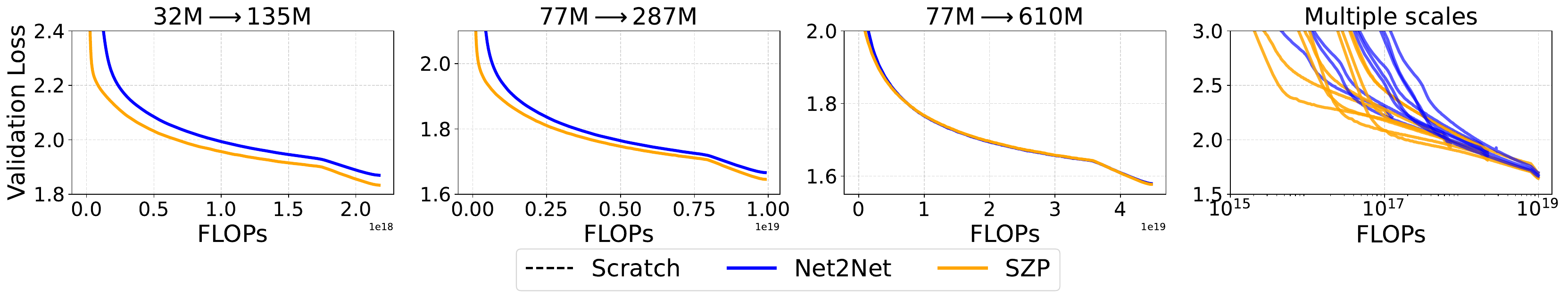}
    \end{tabular}
    \caption{Warmstarting from a \textit{base}\textrightarrow\textit{target} model size, trained for $20$ \tkpm.
    \textit{Top row}: comparing \scratch{} to \ours{} up to $1.2$B parameters, showing that \ours{} converges faster and achieves better final loss, with gains varying across growth factors $g$ (rightmost).
    \textit{Bottom row}: comparing \nton{} to \ours{} (both \ws{} methods explored here) for up to $610$M parameters, showing that \ours{} consistently outperforms the \fp{}-based baseline.}
    \label{fig:n2n-snp-main}
\end{figure}

\paragraph{Empirical Performance of \ours{}.} 
We report performance scaling for \ours{} up to $1.2$B parameters in~\Cref{fig:n2n-snp-main} (top), including a single-checkpoint downstream sanity check in Appendix~\ref{app:downstream-metrics}.
Both rows show that \ours{} converges faster than \scratch{} and \nton{} (for \hc{} results see Appendix~\ref{app:hypercloning})).
Most notably, the relative performance gains of the \ws{} operations (\nton{} and \ours{}) diminish as the growth factor $g$ grows.
\Cref{fig:n2n-snp-main} (rightmost, both rows) shows all the learning curves trained for $10,~20,~\text{and}~30$~\tkpm{}s, for different base-target pairings, with identical learning rate schedules, confirming that faster convergence holds across scales for both comparisons.
The difference in final loss is more subtle and is discussed more in later sections, where we fit scaling laws and systematically characterize the regimes in which \ws{} outperforms training from \scratch{}, and \ours{} outperforms \nton{}.

The variance in performance difference across growth factors is a central observation of this work.
The base model used to derive grown weights is typically well-trained, with its learned function occupying a subspace of the weight space, and exploiting this subspace is where \ws{}'s efficiency gains originate (signal preservation).
For \FP{} methods, which typically rely on \textit{clone} operations, this subspace can be hard to escape due to inherent symmetries in the grown weights~\citep{yu-arxiv26a}.
\Cref{fig:post-hoc-inter-main} (right) (see, Appendix~\ref{app:interpretability}) reflects this in the lower effective rank of \nton{} compared to \ours{}.
Decomposing weight movement into radial (along the initialization) and angular (orthogonal) components, we find that \ours{}'s attention layers show both rotation and scaling of the weights at initialization, while \nton{}'s radial component remains much more stable and weight adjustment is dominated by rotations than scaling the initialization vector.
The practical impact of this may show up in too well-trained base models saturating \ws{} performance~\citep{liew-arxiv25a} or growing a base checkpoint too large such that gains diminish. 
In the next section, we observe and discuss this in more detail.

\begin{figure}[htbp]
    \centering
    \begin{tabular}{cc}
       \includegraphics[width=0.48\linewidth]{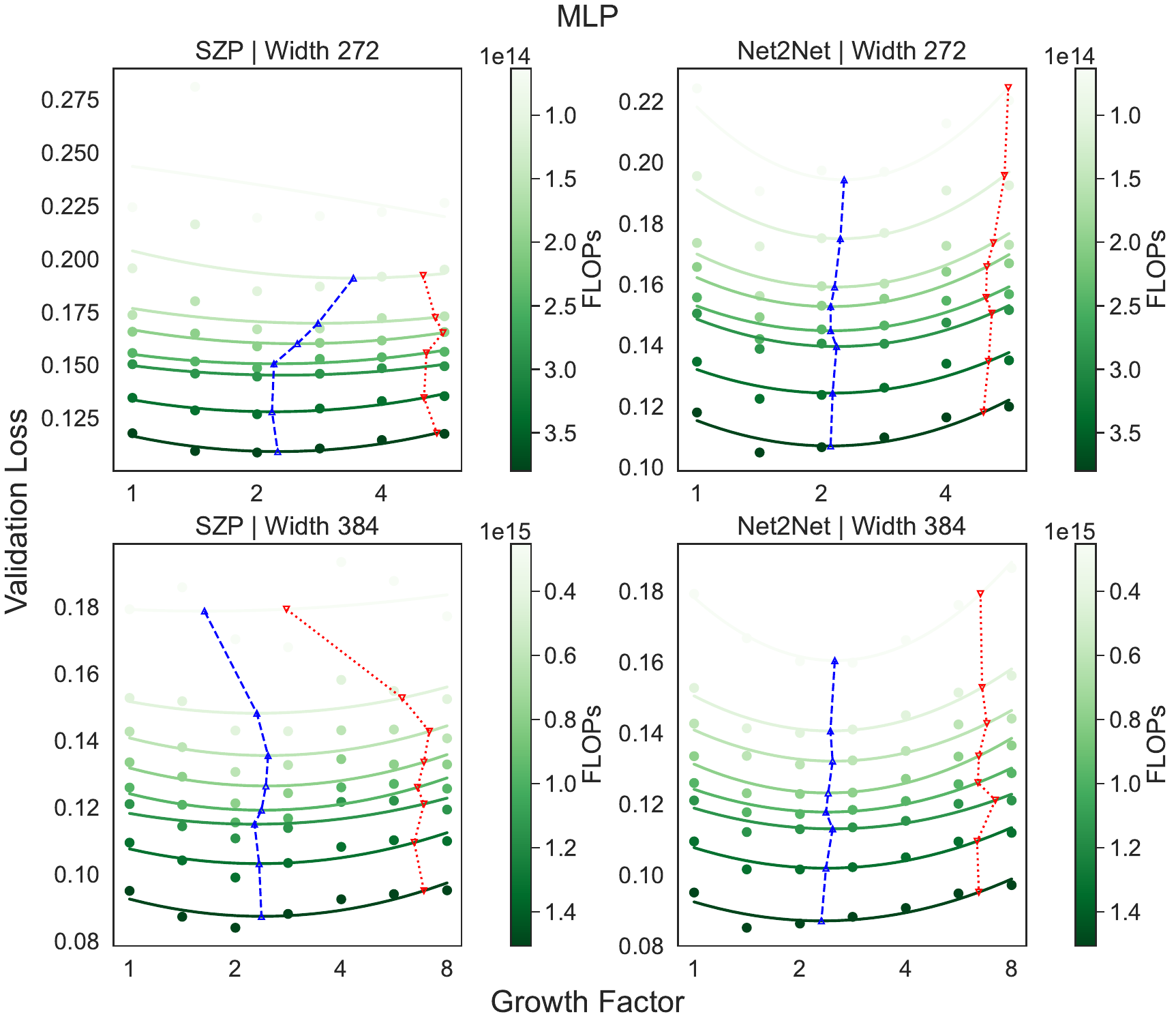} &
       \includegraphics[width=0.48\linewidth]{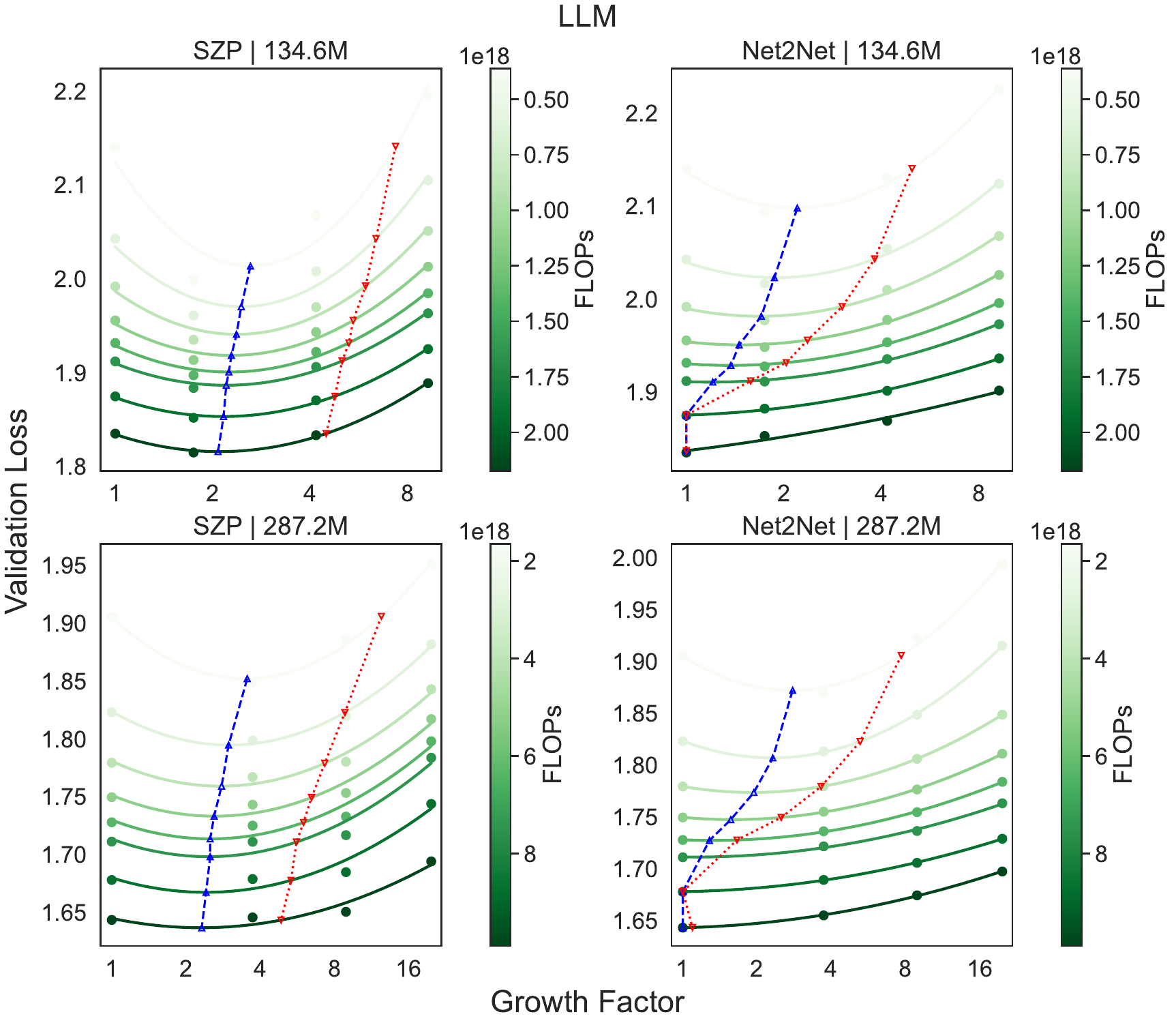} 
    \end{tabular}
    \caption{IsoFLOPs on Language Models and MLPs for 20 \tkpm.
    {\color{blue}Blue lines} (the line connecting the trough of the IsoFLOPs) mark the optimal growth factor,~$\gopt{}$, at each IsoFLOP line, while {\color{red}red lines} (the vertical line right of the $\gopt$ line) mark the $\gub$, the upper bound on the growth factor at which \ws{} matches the loss of training from \scratch{} ($g=1$).
    These lines are obtained from a quadratic fit to the validation loss rather than directly from the loss values.
    Additional results for more target sizes and budgets can be found in Appendix~\ref{app:more_isoflops}.
    }
    \label{fig:isoflops-main}
\end{figure}
\paragraph{The Bounded Effective Growth Factor.}
For practitioners with a fixed base checkpoint seeking to scale up (as scoped in \Cref{sec:introduction}), a key question is how much the model can grow while preserving \ws{}'s efficiency gains over \scratch{}.
Prior work~\citep{du-neurips24a} shows that \ws{} performance depends on the growth factor $g$, defined as the ratio of target to base model parameters, and characterizes this via the optimal growth factor.
This optimal factor, which we term $\gopt$, is estimated by fitting quadratic models to IsoFLOP losses and minimizing them.
$\gopt$ is meant to be an \textit{optimal} growth factor such that it maximizes the relative gains of \ws{} over training from \scratch{}.
We argue this optimization-based characterization is hard to support from sparse evaluations, since reliably identifying $\gopt$ requires substantially denser measurements, especially for $g\in(1,\gopt{})$.
Instead, we propose reporting $\gub$, which we observe as the largest growth factor under which \ws{} still improves over \scratch{}'s final validation loss.
This can be extracted by drawing a horizontal line from $g=1$ for each IsoFLOP line, cutting the parabola at a $g>1$.
We posit that $\gub$ offers a more flexible and practical recommendation for \ws{} in practice, and assess relative gains of different \ws{} operations.

While such a bound is implicit in prior results~\cite{du-neurips24a}, it has not been reported as a quantity of interest.
In~\Cref{fig:isoflops-main}, we reproduce these results for \ours{} and \nton{} in the LLM setting (MLP included as verification), modeling IsoFLOP lines with quadratic fits across growth factors ($g=1$ denotes \scratch{}). 
Both $\gopt$ (blue) and $\gub$ (red) are read off the fit.
Bootstrap uncertainty estimates for both quantities are provided in Appendix~\ref{app:more_isoflops}.
$\gub$ is observed across all settings. 
The exception is \nton{} in the LLM setting, where $\gub=1$ (the \scratch{} run) at the final IsoFLOPs eventually beats \nton{} by the end of the scheduled budget (when including the learning rate decay).
This is consistent with~\cite{du-neurips24a}\footnote{\citet{du-neurips24a} reports that width-only scaling with \fp{} underperforms, while their IsoFLOP analysis on growth factors for width-only scaling also exhibits a $\gub$ bound.} and highlights the shortcomings of function-preservation methods on more complex architectures (refer to Appendix~\ref{app:net2net}).
For complete IsoFLOPs results and an additional analysis accounting for base-model training compute, refer to Appendix~\ref{app:more_isoflops}.

One interpretation is that \WS{} places the target model in a region of the loss landscape shaped by the base model's learned representations. 
For small $g$, this initialization provides a favorable basin for efficient convergence. 
As $g$ grows, the base model occupies a diminishing fraction of the target architecture, and the initialization increasingly resembles a random one. 
Beyond $\gub$, the constraints imposed by the inherited structure outweigh its benefits, anchoring the warmstarted model in a suboptimal region.
Consistent with this, our mechanistic analysis (\Cref{fig:interpret_over_g}) shows that warmstarted models at higher $g$ increasingly resemble \scratch{} across interpretability metrics, suggesting gradual dilution of the base model's structure. 
The stability of $\gub$ across model scales further supports this.
The rate at which warmstarted representations converge to \scratch{}-like behavior depends primarily on the architectural ratio $g$, and not on absolute model size.

\textit{In summary}, we observe a bound on the effective growth factor $\gub$, consistent with~\citet{du-neurips24a}, and argue that identifying and reporting it is crucial for practical adoption of \ws{}.
When budget does not permit investigating $\gub$, we recommend $g=2$ as a conservative default.
Much of the \ws{} literature restricts to $2\times$ growth~\citep{samragh-arxiv24a,ma-arxiv26a}, leaving $\gub{}$ undiscussed.
For fair comparison of \ws{} operations, $\gub{}$ should be discussed and, where possible, investigated.

\begin{figure*}[htbp]
    \centering
    {
    \begin{tabular}{cc} 
        \includegraphics[width=0.47\columnwidth]{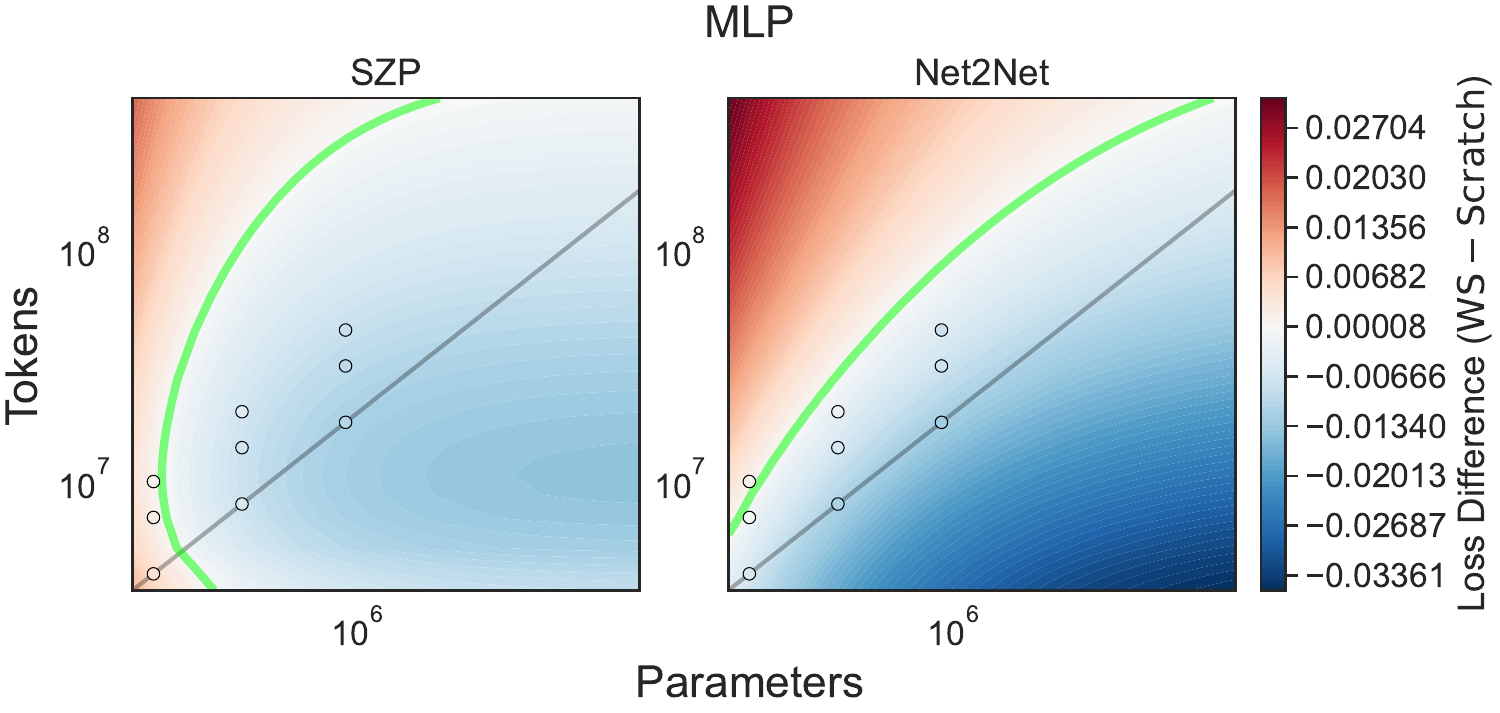} &
        \includegraphics[width=0.47\columnwidth]{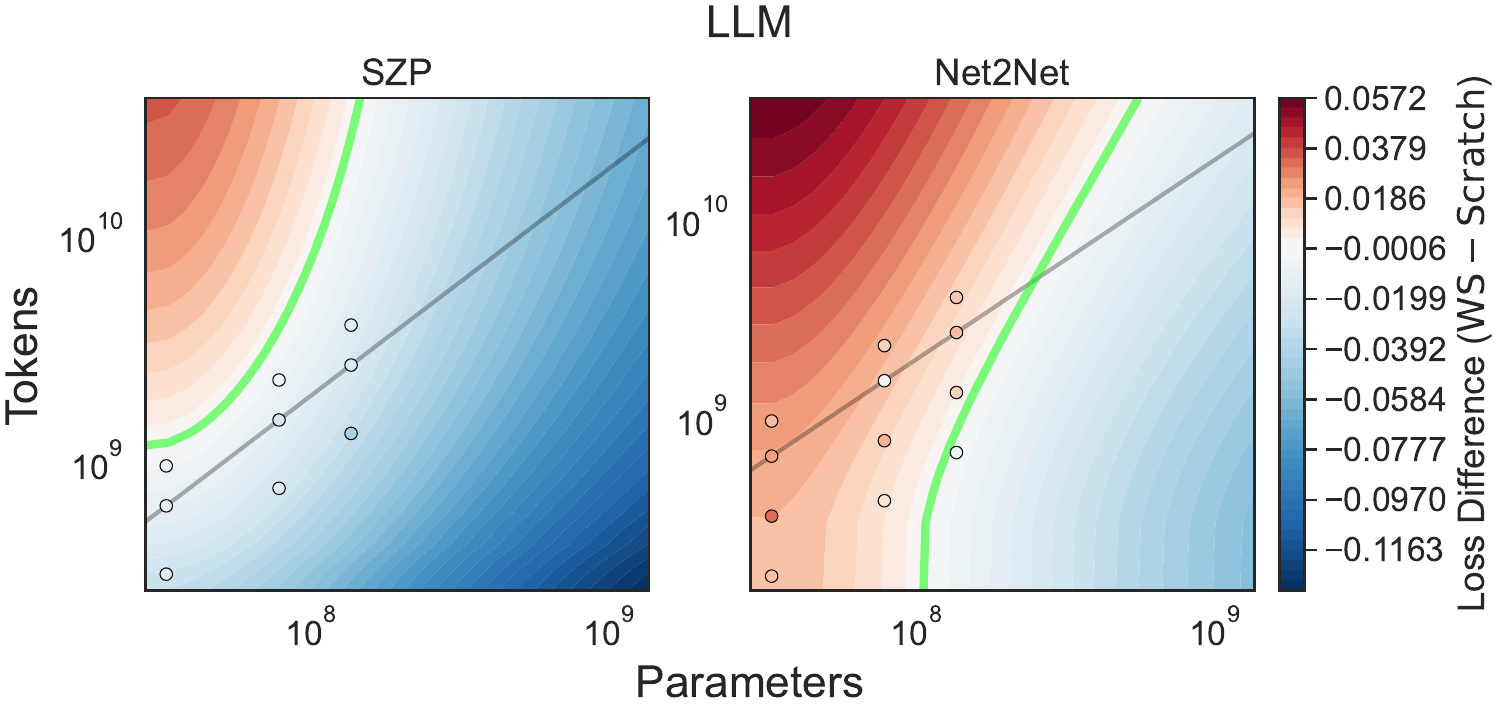} 
         \\
    \end{tabular}
    }
    \caption{
    Isolines for \textit{loss difference} showing where \WS{} achieves lower loss than training from \scratch{} for growth factor $g\approx2$ ({\color{blue}blue regions} indicate \WS{} is better; {\color{red}red} indicates \scratch{} is better; the {\color{green}green line} marks the boundary where both perform equally).
    Given a fixed model size, the more the \tkpm{}, the more preferred is training from \scratch{} over \WS{}.
    The region where \ours{} is effective is larger than that of Net2Net, enabling a more complete comparison between growth methods.
    }
    \label{fig:llm_N_D_ratio}
\end{figure*}

\paragraph{Effective Warmstarting Across Scale.}
Scaling a language model involves growing both the model and the data budget. 
As discussed, diminishing gains are expected from \ws{}, so studying its effectiveness at a single token budget and sparse model sizes provides only a limited, possibly misleading, signal. 
While~\citet{liew-arxiv25a} identified that overtraining base models diminishes \ws{} gains, we highlight the effect of $\gub$ on parameter scaling.
Since we assume the base model's training budget is unknown, we focus on the practical requirement of specifying the target budget.

We subset our collected data for $g \approx 2$ (\Cref{fig:isoflops-main}) and follow Approach 3 from~\citet{hoffmann-neurips22a} to fit the power-law parametric form predicting loss as a function of parameters $N$ and tokens $D$,
\begin{equation*}
L(N, D) = E + \frac{A}{N^{\alpha}} + \frac{B}{D^{\beta}},
\end{equation*}
where $E, A, \alpha, B, \beta$ are fit by optimizing the Huber loss ($\delta=10^{-3}$) on $\log L$ with L-BFGS-B. We fit scaling laws for \scratch{}, \ours{}, and \nton{} in both LLM and MLP settings. 
Appendix~\ref{app:scaling-fit} reports the parameters and $R^2$ values, all above $0.99$, confirming fit quality.
In~\cref{fig:llm_N_D_ratio}, we visualize the \emph{difference} in modeled loss between \ws{} methods and \scratch{} fits as \textit{Isolines} (contour lines of equal loss difference), marking the zero-difference contour in green as the boundary for our analysis. 
First, the existence of this boundary across MLPs and LLMs indicates regimes where training from \scratch{} can be preferred over \ws{}. 
Second, the boundary's position relative to the Chinchilla-optimal \tkpm{} front (solid black line) suggests that training for fewer tokens/parameter is more likely to retain \ws{} gains. 
Importantly, this bound provides practical guidance, subject to the quality of the fits.

These scaling-law fits and \textit{Iso-Loss Difference} contours provide a wider picture that corroborates our initial hypothesis on the necessity of \FP{} for \ws{}. 
In the MLP setup, \nton{} provides a clear front of \ws{} gains with stronger final performance than \scratch{}.
This is an expected result, as \FP{} was designed for MLPs. 
However, \ours{} provides a larger range of \ws{} efficacy, with the loss-difference front pushed further from the Chinchilla-optimal front. 
As a contrasting case, for LLMs we observe that \nton{} shows only a small region of predicted gains over \scratch{}, consistent with the trends in \Cref{fig:isoflops-main}.
For LLMs, \ours{} offers a reasonable front, indicating scaling trends for when \ws{} becomes worse than \scratch{} training.

\textit{In summary}, shorter training budgets ($\le 20~\tkpm{}$) are more likely to yield \ws{} gains over \scratch{}.
For fair comparison of \ws{} operations, the scaling pattern must be assessed and the front beyond which \ws{} loses its advantage characterized, yielding practical guidance.

\section{Conclusion}\label{sec:conclusion}

We have characterized the upper bound on the growth factor within which \WS{} (\ws{}) remains computationally advantageous over training from \scratch{}. 
Beyond $\gub$, efficiency gains diminish significantly. 
Our analysis reveals that this limit is fundamentally tied to the training budget. 
We distill our contributions into a set of principled guidelines for the community:

\begin{tcolorbox}[
    enhanced,
    colback=white,
    colframe=white,
    boxrule=0pt,
    sharp corners,
    left=8pt,
    right=0pt,
    top=2pt,
    bottom=2pt,
    borderline west={2pt}{0pt}{gray!65}
]
{\setlength{\parindent}{0pt}%
\begin{list}{$\bullet$}{%
    \setlength{\leftmargin}{1.2em}%
    \setlength{\labelsep}{0.5em}%
    \setlength{\itemsep}{1pt}%
    \setlength{\parsep}{0pt}%
    \setlength{\parskip}{0pt}%
    \setlength{\topsep}{1pt}%
    \setlength{\partopsep}{0pt}%
}
    \item \textbf{Model Scaling:} If only one model run is feasible, $g=2$ is the most reliable scaling for \ws{}.
    \item \textbf{Efficiency Window:} \ws{} should be prioritized over \scratch{} when training at or below compute-optimal scaling ($20$~\tkpm{}).
    \item \textbf{LR Transfer:} $\mu\text{P}$ serves as a principled strategy for hyperparameter transfer in \ws{}, particularly when tuning budgets for larger scales are constrained.
    \item \textbf{Standardized Reporting:} We propose future \ws{} literature explicitly report the empirical $\gub$ wherever possible, and that comparisons account for scaling patterns of diminishing \ws{} gains.
\end{list}
}
\end{tcolorbox}

We show that \ours~(\emph{Shrink-Zero-Perturb}), which augments zero-padding with a \snp{} strategy, is architecture-agnostic and more broadly applicable than function-preserving methods.
\ours{} is easier to interpret, control, and study across architectures and setups, and our work here offers future \ws{} literature both a framework to assess limits and make fairer comparisons of \ws{} operations.

\paragraph{Limitations.}
Our empirical study is constrained by the computational costs of language model training and the need for controlled comparisons.
To reflect realistic scenarios where only model checkpoints are available, we treat the base model as a black box regarding its training history: \ours{} needs only the weight checkpoint, not optimizer states or schedule information~\citep{shen-pmlr2022}.
While this distinguishes our work from studies that assume full access to base training compute~\citep{liew-arxiv25a}, it limits our ability to analyze the joint compute-optimal frontier across base size, target size, and tokens invested in both training phases.
However, our findings at sub-$1$B scales align with trends reported at $1$B+ scales in prior work, suggesting transferability to larger regimes.
Additionally, exhaustive hyperparameter sweeps across all growth operators remain computationally prohibitive at the language model scale.
Our comparison focuses on \ours{} vs. \nton{} to establish a representative baseline, with the understanding that \ours{} can serve as a modular alternative for width-scaling methods extending \nton{}.
Especially, combining with depth-scaling~\citep{du-neurips24a} for jointly growing both depth and width.

\paragraph{Future Directions.}
Several promising avenues emerge from this work.
First, benchmarking \ours{} across diverse architectures and data modalities would validate its generality beyond language models.
Second, understanding the individual contributions of shrinking and perturbation within \snp{} could reveal whether these components can be predictively scaled using fixed rules or scaling law fits, and extending to \ws{} of optimizers too.
Third, given the existence of upper bounds on growth factors, exploring progressive growing and optimal growth schedules becomes a natural next step.
Fourth, extending the scaling law to jointly model loss as a function of parameters, tokens, and growth factor would remove the need to fix the growth factor across experiments, enabling predictions for arbitrary growth factors from a single unified fit.
More broadly, as specialized model checkpoints become increasingly available, strengthening the practical impact of warmstarting methods will be critical for improving resource efficiency across the field.

\section*{Acknowledgements}

We thank Tarek Abou Chakra and Samir Garibov for their code support in the initial days of the project. We thank André Biedenkapp for feedback on our paper.

\textbf{FH} acknowledges the financial support of the Hector Foundation.
\textbf{NM}, \textbf{JH}, \textbf{HR}, \textbf{FH} acknowledge funding by 
the state of Baden-W\"{u}rttemberg through bwHPC, the German Research Foundation (DFG) through grant numbers INST 39/963-1 FUGG and 417962828, and the European Union (via ERC Consolidator Grant Deep Learning 2.0, grant no.~101045765). 
Views and opinions expressed are however those of the author(s) only and do not necessarily reflect those of the European Union or the European Research Council. 
Neither the European Union nor the granting authority can be held responsible for them.
This research was also funded by the Deutsche Forschungsgemeinschaft (DFG, German Research Foundation) under grant number 539134284, through EFRE (FEIH\_2698644) and the state of Baden-Württemberg.
\textbf{NM} is supported by the Konrad Zuse School of Excellence in Learning and Intelligent Systems (ELIZA) through the DAAD programme Konrad Zuse Schools of Excellence in Artificial Intelligence, sponsored by the Federal Ministry of Education and Research.
\textbf{JH} and \textbf{FH} acknowledge funding by the Deutsche Forschungsgemeinschaft (DFG, German Research Foundation) under SFB 1597 (SmallData), grant number 499552394.
\textbf{MJ} and \textbf{JG} acknowledge funding by The Carl Zeiss Foundation through the research network "Responsive and Scalable Learning for Robots Assisting Humans" (ReScaLe) of the University of Freiburg.
\textbf{AK} acknowledges support from EC under the grant No. 101195233 (OpenEuroLLM).
\begin{center}
\includegraphics[width=0.25\textwidth]{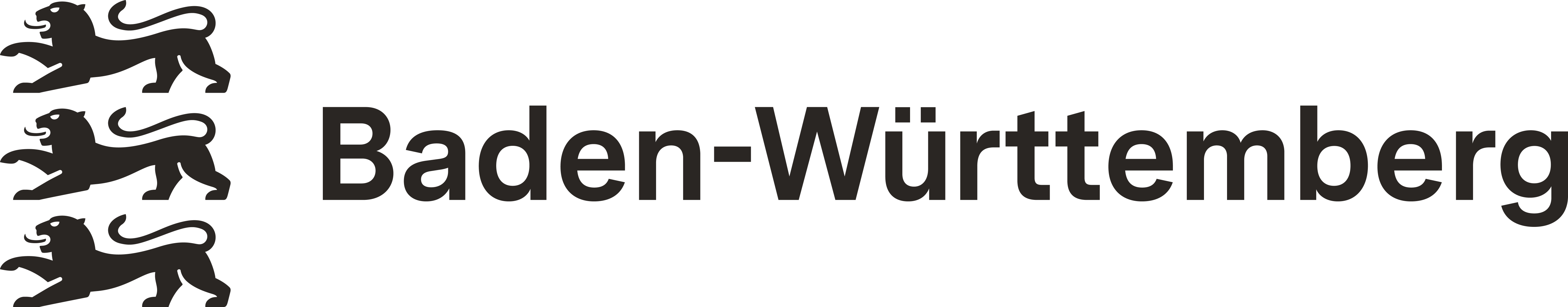} \quad
\includegraphics[width=0.3\textwidth]{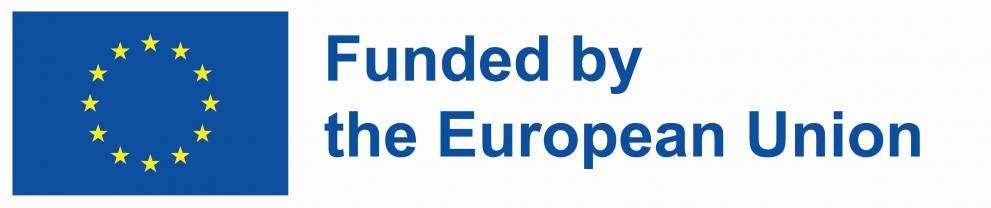} \quad
\includegraphics[width=0.12\textwidth]{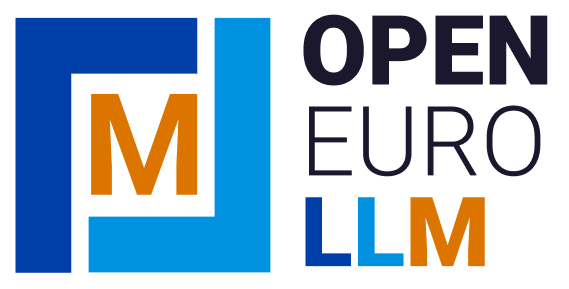}
\end{center}

\bibliographystyle{plainnat}
\bibliography{bib/lib,bib/local,bib/proc,bib/strings}

\newpage

\appendix

\appendix
\onecolumn

\section{Supporting Related Work}\label{app:related}

\textbf{Scaled parameterizations} refer to a set of rules that determine how certain parameters can be scaled with respect to one or more scaling dimensions~\citep{everett-icml24a}. 
This family of parameterizations describes scaling factors for the weights, the learning rate, and the standard deviation of the initialization.
Each of these scaling variables can be defined separately for the input, hidden, and output layers (see Tables 1, 2, 3\footnote{and the $\mup{}$ Github page: \url{https://github.com/microsoft/mup/blob/main/README.md}} in~\citet{yang-icml21a}).
The maximal update parameterization or $\mup{}$ is one such parameterization that showed its scaling rules to allow for stable training of larger networks without requiring re-tuning of the variables (such as learning rate) being scaled~\citep{yang-neurips21a}, i.e., feature learning does occur at infinite-width scaling~\citep{yang-icml21a}, extending the Neural Tangent Kernel~\citep{jacot-neurips18a} theory demonstrating lazy-training (fixed features) when the network is scaled to infinite-width.
Follow-up works exploited the symmetry in these parameterizations to provide alternative scaling rules~\citep{blake-iclr25a}, extended to meta learning~\citep{therien-arxiv24a}, and applied to newer optimizers~\citep{essentialai-arxiv25a}.
Generally, in practice, a network where the learning rate is adequately tuned on a small-width network can be scaled using $\mup{}$ to be used as the learning rate for a larger width (theoretically, $\infty$-width), and still exhibit stable training dynamics (no blowup of L1 norms of pre-activations or output logits).
While $\mu$P does not inherently predict loss values, its ability to preserve feature learning remains effective across diverse empirical settings~\citep{vyas-neurips23a}.
Consequently, hyperparameter transfer is well-aligned with the predictable loss trajectories observed in language model training~\citep{dey-arxiv24a,bjorck-iclr25a}.

\textbf{Scaling laws} formalize these trajectories by deriving parametric power-law fits of optimal loss against compute, dataset size, or parameter count~\citep{kaplan-arxiv20a,hoffmann-neurips22a,sorscher-neurips22a,alabdulmohsin-neurips23a,caballero-iclr23a}.
Given a pretraining dataset and model architecture type, multiple models are trained varying the number of tokens seen, the model hyperparameters (including batch size), and the model sizes (width-depth).
The resulting Pareto front characterizes the loss-compute trade-off, which is approximately linear on log-log axes and is commonly modeled with a power-law fit.
This fit can then be used to predict the tokens required for compute-optimal loss given a fixed model, or the model size required for compute-optimal loss given a fixed token set to process.
Chinchilla~\citep{hoffmann-neurips22a} notably showed that this fit is sensitive to the hyperparameter choices, e.g., the length of cosine decay in the learning rate schedule.
Subsequent works have looked at such a role of hyperparameters when scaling models~\citep{porian-icml24a}, and also the suitability of certain hyperparameters to allow for compute reuse while scaling~\citep{hagele-neurips24a}.
Often, the role of \textit{optimal} hyperparameters is highlighted as crucial but inadequately reported.
Hyperparameter transfer methods become a key ingredient of a scaling recipe, and also influence novel observations for the scaling regime, such as scaling collapse~\citep{qiu-icml25a}.
\citet{bergsma-arxiv25a} look at how $\mup$, collapsed scaling curves, and novel parametric forms for scaling relationships together can be leveraged for early stopping learning curves.

\section{Synthetic Regression Benchmark}
\label{app:synthetic-regression}

We use the synthetic regression benchmark in which the target function is constructed to have a power-law Fourier spectrum~\citep{qiu-icml25a}. 
Following their compute accounting, training progress is measured by the number of \emph{processed training examples}; throughout, we refer to one regression example as one ``token''. 
To enable extensive hyperparameter optimization, we scale down the models while keeping the task and accounting unchanged. 
We fix the MLP depth to $L=3$ hidden layers and scale width ($n_{\text{embd}}$), resulting in the model sizes in Table~\ref{tab:scale-table}. 

\paragraph{Data Generation.}
Each regression token consists of an input vector $x \in \mathbb{R}^{d}$ sampled from a fixed distribution (uniform on a bounded domain in the standard setup), and a scalar target
\begin{equation}
y = f^\star(x) + \epsilon,
\label{eq:synthetic-regression-data}
\end{equation}
where $\epsilon$ is i.i.d.\ noise and $f^\star$ is a random function whose Fourier coefficients decay as a power law. 
Concretely, $f^\star$ is constructed by sampling Fourier amplitudes with
\begin{equation}
\mathbb{E}\big[|\hat{f}(\omega)|^2\big] \propto \|\omega\|^{-\alpha},
\label{eq:fourier-powerlaw}
\end{equation}
so that the target spectrum follows a controlled slope $\alpha$. 

\paragraph{Model.}
We train width-scaled MLPs of depth $L=3$. Let $h^{(0)} = x$ and for $\ell = 1,\dots,L$,
\begin{equation}
h^{(\ell)} = \phi\!\left(W^{(\ell)} h^{(\ell-1)} + b^{(\ell)}\right),
\qquad 
\hat{y} = W^{(L+1)} h^{(L)} + b^{(L+1)}.
\label{eq:mlp-definition}
\end{equation}
We scale the hidden dimension $n_{\text{embd}}$ while keeping depth fixed.

Although architecturally simple, these MLPs follow the parameterization and normalization conventions of Transformer feedforward blocks, with attention removed; thus they are not treated as generic fully-connected networks but as ``Transformer MLP-only'' models to preserve scaling-recipe behavior under the same compute accounting.

\begin{table}[t]
    \centering
    \caption{
    Synthetic MLP model scales used in the regression experiments.
    All models have $L=3$ hidden layers; we scale only the hidden width $n_{\mathrm{embd}}$.
    Parameter counts are rounded to one decimal place.
    }
    \label{tab:scale-table}
    \small
    \setlength{\tabcolsep}{8pt}
    \renewcommand{\arraystretch}{1.05}
    \begin{tabular}{lc}
    \toprule
    \textbf{Width ($n_{\mathrm{embd}}$}) & \textbf{Params (M)} \\
    \midrule
    $48$  & $0.1$ \\
    $68$  & $0.1$ \\
    $96$  & $0.2$ \\
    $136$ & $0.4$ \\
    $192$ & $0.9$ \\
    $272$ & $1.8$ \\
    $384$ & 3.5 \\
    \bottomrule
    \end{tabular}
\end{table}

\subsection{Training Horizon via \tkpm{}}
\label{app:regression_horizon}

For a model with $P$ parameters, we set the training horizon using a \tkpm{} budget $\tau$ and process a total of
\[
T \;=\; \tau \cdot P
\]
training examples (tokens). Given batch size $B$, this corresponds to $T/B$ optimizer steps. We run experiments for $\tau \in \{20, 35, 50\}$.

\citet{qiu-icml25a} measure compute in processed examples and, for some comparisons, estimate a compute-optimal horizon as a function of model size (via a loss--compute frontier under a reference schedule) and then reuse that horizon when comparing schedules. In our setting, we fix $\tau$ explicitly (rather than re-estimating compute-optimal horizons) so that large, matched hyperparameter sweeps remain feasible and comparable across scales.

\subsection{Hyperparameter Search Spaces}
\label{app:regression_hpspaces}

We use a deterministic Cartesian-product grid search with a fixed seed (no adaptive tuning). To keep the overall cost manageable, we distinguish:
(1) a \emph{base grid} evaluated at the smallest width, used to identify strong configurations and for transfer experiments; and
(2) a \emph{main grid} used for the majority of runs across target widths.

\paragraph{Grid Definitions.}
Table~\ref{tab:hyperparameter-search-space} lists both grids. The base grid expands the learning-rate and batch-size options, while the main grid keeps a smaller but representative range. In all cases the reported distributions (e.g., Fig.~\ref{fig:violin-hp}) are over the \emph{full grid at the target width}.

\begin{table}[t]
\centering
\caption{Hyperparameter search spaces for the regression benchmark. The base grid is evaluated at the smallest width ($48$) and used for transfer; the main grid is used for the primary sweeps across target widths.}
\begin{tabular}{ll}
\toprule
\textbf{Hyperparameter} & \textbf{Values} \\
\midrule
\multicolumn{2}{l}{\textbf{Base grid (evaluated at width = $48$)}} \\
\midrule
Learning rate & \{6e-4, 8e-4, 1e-3, 1.4e-3, 1.8e-3, 3e-3, 4e-3, 5e-3\} \\
Weight decay & \{0, 1e-5, 1e-2\} \\
Batch size & \{256, 512, 1024, 2048, 4096\} \\
Warmup fraction & \{0.01, 0.03, 0.05\} \\
Cooldown fraction & \{0, 0.2\} \\
\midrule
\multicolumn{2}{l}{\textbf{Main grid (used for target widths)}} \\
\midrule
Learning rate & \{6e-4, 1e-3, 3e-3, 4e-3\} \\
Weight decay & \{0, 1e-5, 1e-2\} \\
Batch size & \{256, 512, 1024\} \\
Warmup fraction & \{0.01, 0.03, 0.05\} \\
Cooldown fraction & \{0, 0.2\} \\
\bottomrule
\end{tabular}
\label{tab:hyperparameter-search-space}
\end{table}

\paragraph{Grid Sizes.}
The base grid contains $8 \cdot 3 \cdot 5 \cdot 3 \cdot 2 = 720$ configurations.
The main grid contains $4 \cdot 3 \cdot 3 \cdot 3 \cdot 2 = 216$ configurations.
Unless stated otherwise, each experimental setting uses the full main grid (216 evaluations).

\subsection{Hyperparameter Importance}
\label{app:regression_hp_importance}

We analyze hyperparameter importance using the results of our deterministic grid searches: the reduced grid is evaluated at target scales, while the full grid is evaluated at the base scale (cf.\ Table~\ref{tab:hyperparameter-search-space}). For each experimental setting, we treat the best validation loss achieved during training (best-so-far over processed examples) as the objective and compute hyperparameter importances using Optuna's fANOVA~\citep{hooker-jcgs07a} implementation.

Across all settings, learning rate is consistently the dominant contributor to performance variation, followed by the learning-rate schedule shape (primarily cooldown fraction, and to a lesser extent warmup fraction). In contrast, weight decay receives near-zero importance in all experiments. We attribute this to the small-model regime in our scaled-down benchmark (Table~\ref{tab:scale-table}), where explicit $\ell_2$ regularization has little effect compared to optimization and schedule choices. Consequently, while we keep weight decay in the search space for completeness and comparability, we do not emphasize it when interpreting transfer behavior and sensitivity analyses.

\begin{figure}[htbp]
    \centering
    \includegraphics[width=0.5\linewidth]{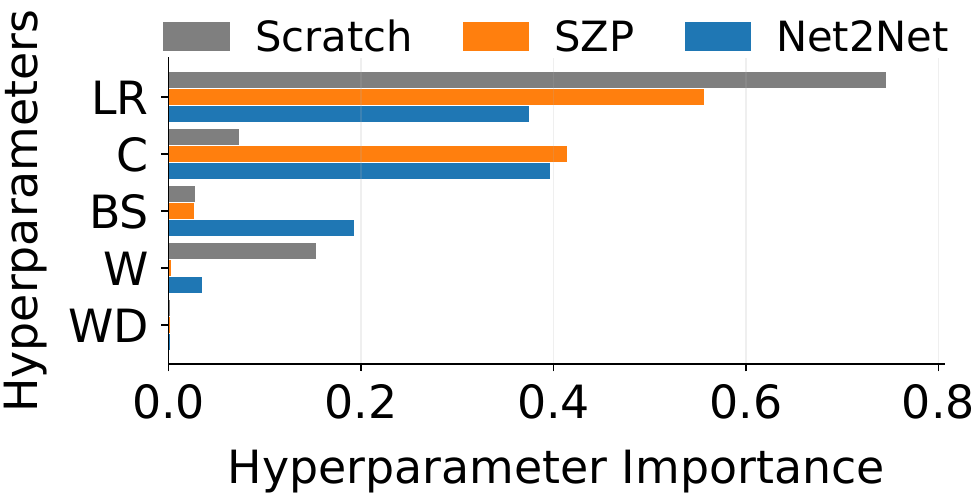}
    \caption{Hyperparameter importance (fANOVA via Optuna), aggregated across all target widths, computed on the main-grid search space (Table~\ref{tab:hyperparameter-search-space}). We report importances separately for \scratch{}, \ours{} warmstarting, and \nton{} warmstarting. Abbreviations: LR = learning rate, WD = weight decay, BS = batch size, W = warmup fraction, C = cooldown fraction.}
    \label{fig:fanova}
\end{figure}

\subsection{Details on $\muP$}
\label{app:mup-details}

Changing the width of a neural network changes not only its parameter count, but also the effective scale of activations, gradients, logits, and parameter updates.
Therefore, hyperparameters tuned at one width need not transfer reliably to another.
This creates a practical confounder for \ws{} experiments: if each target width is retuned independently, improvements may come from better target-scale hyperparameters rather than from the growth operator itself.

$\muP$~\citep{yang-neurips21a} addresses this issue by prescribing width-dependent initialization and learning-rate scalings.
Its goal is to keep models of different widths in a comparable feature-learning regime, so that the effect of parameter updates on activations remains stable as width changes.
Equivalently, $\mup{}$ can be viewed as a hyperparameter-transfer rule from a small base model to a larger target model.

Given a base-width configuration
\[
h_{n_0}
=
(\eta, \lambda_{\mathrm{wd}}, B, w_{\mathrm{warmup}}, w_{\mathrm{cooldown}}, \ldots),
\]
we obtain the target-width configuration by
\[
h_n = T_{\mup}(h_{n_0}; n_0, n),
\]
where $T_{\mup{}}$ applies parameter-type-specific scaling rules to initialization and optimizer hyperparameters.
The main rules are summarized in \cref{tab:mup-transfer-rules}.
The key point is that different tensors can have different width dependencies: input weights, hidden weights, readout weights, embeddings, normalization parameters, and attention-logit scaling are not all treated in the same way.
In practice, we specify base shapes for the model, identify which dimensions grow with width, and apply the corresponding $\mup{}$ optimizer rules.

\begin{figure}[htbp]
    \centering
    \includegraphics[width=0.95\linewidth]{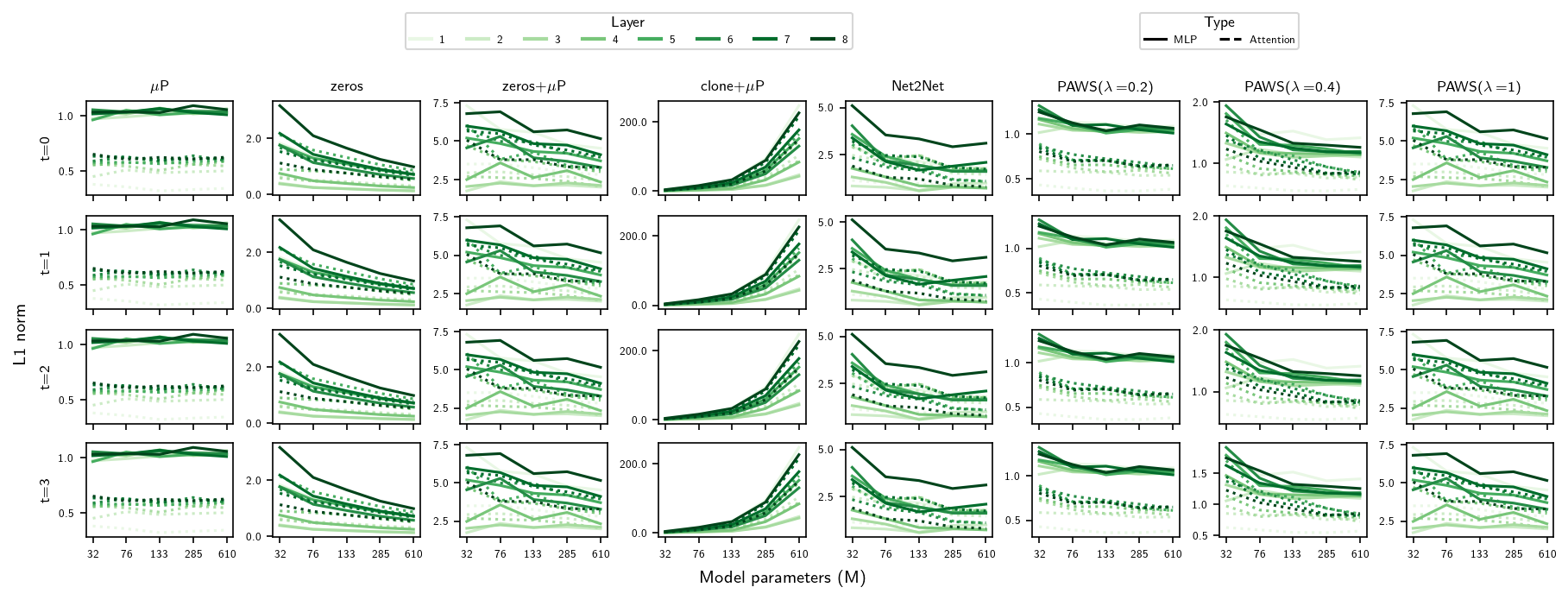}
    \caption{L1 norm of the layer activations across model sizes, for $\mup{}$, \ours{}, \nton{} and the other naive baselines discused in~\Cref{sec:methodology}, for $t=\{1,\ldots,4\}$.
    The expectation is for each line to resemble the $\mup$ behaviour.
    However, in practice, and in expectation of the theory, the norms \emph{not} blowing up with increasing width is adequate for stable hyperparameter transfer.
    We find that for most cases at larger widths, the activations become more stable and constant in width, especially for \ours{} (with $\shrinkhp=0.2$).
    Only for \texttt{Clone} as $\growth$ do we observe the prohibitive instability with increasing width.
    }
    \label{fig:l1-check}
\end{figure}

We use $\mup{}$ only as a practical mechanism for hyperparameter transfer.
In the synthetic MLP benchmark, we can compare it directly against dense target-width tuning and against \emph{static transfer}, where the best base-width hyperparameters are reused unchanged.
In \cref{fig:violin-hp}, stars denote $\mup$ transfer and open circles denote static transfer.
For language modeling, full target-scale sweeps are too expensive, so we tune at the smallest relevant scale and transfer with $\mup{}$ throughout.
In~\Cref{fig:l1-check}, we verify the baselines and algorithms used in the empirical experiments in this work based on the scaling of the activations with increasing width as recommended by~\citet{yang-neurips21a}.

\begin{table}[t]
\centering
\caption{
Compact summary of the $\mup$ transfer rules used in our experiments.
Here, $n$ denotes the width-like dimension for tensors whose fan-in/fan-out grows with model width.
For attention logits, the scaling uses the per-head dimension $d_{\mathrm{head}}$, which is fixed to $64$ in our language-model width-scaling setup.
}
\label{tab:mup-transfer-rules}
\small
\setlength{\tabcolsep}{4pt}
\renewcommand{\arraystretch}{1.12}
\begin{tabular}{p{0.32\linewidth}p{0.30\linewidth}p{0.25\linewidth}}
\toprule
\textbf{Group} & \textbf{Forward / init} & \textbf{Adam LR} \\
\midrule
Input / vector-like & $\mathrm{Var}(W)\propto 1/n$ & constant \\
Hidden / matrix-like & $\mathrm{Var}(W)\propto 1/n$ & $\propto 1/n$ \\
Output / readout & multiplier $\propto 1/n$ & constant \\
Attention logits & $q^\top k / d_{\mathrm{head}}$ & -- \\
Scalar-like & constant & constant \\
\bottomrule
\end{tabular}
\end{table}

\subsection{Hyperparameter Transfer across Widths}
\label{app:regression_hp_transfer}

We study whether hyperparameters selected at the smallest width transfer to larger widths. 
For each target scale, we evaluate: (1) the full main grid sweep at the target width, and (2) a single transferred run obtained by mapping the best configuration found at the base scale (width = $48$) to the target width.

\paragraph{Transfer Rules.}
Under $\mu$P, the learning rate selected at the base width is transferred according to the $\mu$P prescription (i.e., it can be reused across widths when training uses the $\mu$P parameterization). In addition, following subsequent work on stabilizing width transfer, we scale batch size with target width as
\[
B_{\text{target}} \;\approx\; B_{\text{base}} \cdot \sqrt{\frac{n_{\text{embd@target}}}{n_{\text{embd@base}}}},
\]
where $n_{\text{embd}}$ denotes model width.

\section{Design Intuition and Ablations for \ours{}}
\label{app:szp-details}

This appendix complements the empirical setup in \cref{sec:empirical-setup} by analyzing the main design choices in \ours{}.
The method combines three simple ingredients: zero-padding, perturbation, and shrinkage.
Zero-padding embeds the trained base model into the larger model, perturbation activates the newly added parameters, and shrinkage reduces the dominance of the copied base-model path.
We first give a simple theoretical intuition for these roles, then provide a mechanistic interpretability analysis of their effect on weight statistics during training, and finally ablate the two nontrivial design choices: the perturbation scale and the shrinkage factor.

\subsection{Theoretical Intuition for \ours{}}
\label{app:szp-theory}

\paragraph{Zero-padding as an Exact Embedding.}
Consider a feed-forward network with hidden states
\[
h_{\ell+1} = \phi(W_\ell h_\ell + b_\ell),
\]
where $\phi(0)=0$.
Let $P(\theta)$ denote the width expansion from width $n$ to width $m>n$ by adding new neurons and setting every parameter touching them to zero.
If we embed each hidden state as
\[
\tilde h_\ell =
\begin{bmatrix}
h_\ell \\
0
\end{bmatrix},
\]
then the widened layer acts as
\[
\tilde h_{\ell+1}
=
\phi\!\left(
\begin{bmatrix}
W_\ell & 0 \\
0 & 0
\end{bmatrix}
\begin{bmatrix}
h_\ell \\
0
\end{bmatrix}
+
\begin{bmatrix}
b_\ell \\
0
\end{bmatrix}
\right)
=
\begin{bmatrix}
\phi(W_\ell h_\ell + b_\ell) \\
0
\end{bmatrix}
=
\begin{bmatrix}
h_{\ell+1} \\
0
\end{bmatrix}.
\]
By induction over layers,
\[
F_{P(\theta)}(x) = f_\theta(x)
\qquad \text{for all } x.
\]
Thus, in this idealized unnormalized setting, zero-padding preserves the pretrained function exactly.
For normalized architectures such as Transformers with LayerNorm or RMSNorm, exact function preservation may additionally require coordinated rescaling of normalization parameters or adjacent weights -- without this, zero-padding should be viewed as an embedding intuition rather than a strict function-preserving operation.

However, this exact embedding is also restrictive.
Let
\[
\mathcal M := \{P(\theta): \theta \in \Theta_{\mathrm{small}}\}
\]
be the embedded small-model manifold.
At points in $\mathcal M$, each newly added neuron has zero activation and zero outgoing weight. 
By the chain rule of backpropagation, because the outgoing weights are exactly zero, the error signal propagated back to these new neurons is zero. 
Consequently, the gradients with respect to all newly added parameters vanish, and under standard gradient descent,
\[
\nabla_{\Theta_{\mathrm{new}}} \mathcal L(P(\theta)) = 0.
\]
Thus, pure zero-padding can keep optimization on the embedded small-model manifold, preventing the wider model from using its additional capacity.

\paragraph{Why Perturbation and Shrinkage Help.}
The \ours{} initialization can be written as
\[
\Theta_0
=
\lambda_{\mathrm{shrink}} P(\theta^\star) + E,
\qquad
0 < \lambda_{\mathrm{shrink}} \le 1,
\]
where $\theta^\star$ is the pretrained base-model solution and $E$ is a perturbation.

For one widened layer, partition old and new coordinates and write the input as
\[
\tilde h =
\begin{bmatrix}
h \\
h_{\mathrm{new}}
\end{bmatrix},
\]
where $h_{\mathrm{new}}=0$ for pure zero-padding and $h_{\mathrm{new}}=O(\|E\|)$ after perturbations from preceding layers.
The following calculation keeps only first-order terms in the perturbation magnitude.
Let the perturbation matrix and bias vector be conformably partitioned into blocks corresponding to the old (o) and new (n) dimensions:$$ E = \begin{bmatrix} E_{oo} & E_{on} \\ E_{no} & E_{nn} \end{bmatrix}, \qquad e = \begin{bmatrix} e_o \\ e_n \end{bmatrix}. $$

The preactivation then takes the form
\[
\tilde z
=
\begin{bmatrix}
\lambda_{\mathrm{shrink}}(Wh+b) \\
0
\end{bmatrix}
+
\begin{bmatrix}
E_{oo}h + e_o \\
E_{no}h + e_n
\end{bmatrix}
+
O(\|E\|^2).
\]

The lower block,
\[
E_{no}h + e_n,
\]
is the first nonzero signal available to the newly added neurons.
If $E=0$, the new coordinates remain inactive.
Thus, perturbation moves the model off the embedded manifold $\mathcal M$ and activates the added degrees of freedom.

Shrinkage plays a complementary role.
It reduces the copied old-path signal from $\|Wh+b\|$ to approximately $\lambda_{\mathrm{shrink}}\|Wh+b\|$, while the new-path signal remains controlled by the perturbation.

The relative influence of the new coordinates is therefore, to first order,
\[
\frac{\|(\tilde z)_{\mathrm{new}}\|}
     {\|(\tilde z)_{\mathrm{old}}\|}
=
\frac{\|E_{no}h + e_n + O(\|E\|^2)\|}
     {\|\lambda_{\mathrm{shrink}}(Wh+b) + E_{oo}h + e_o + O(\|E\|^2)\|}.
\]
When the copied old-path signal dominates the old-block perturbation, i.e.,
\[
\lambda_{\mathrm{shrink}}\|Wh+b\| \gg \|E_{oo}h+e_o\|,
\]
this simplifies to
\[
\frac{\|(\tilde z)_{\mathrm{new}}\|}
     {\|(\tilde z)_{\mathrm{old}}\|}
\approx
\frac{\|E_{no}h + e_n\|}
     {\lambda_{\mathrm{shrink}}\|Wh+b\|}.
\]

Taking $\lambda_{\mathrm{shrink}}<1$ makes it easier for the newly added neurons to compete with the copied subnetwork.
Crucially, a larger relative preactivation ensures that the newly added weights generate correspondingly larger gradients during backpropagation, actively pulling the optimization trajectory away from the restrictive manifold $\mathcal M$. 
In summary, zero-padding preserves the pretrained model, perturbation activates the new dimensions, and shrinkage accelerates adaptation by breaking the optimization dominance of the inherited narrow solution.

\subsection{Interpretability}
\label{app:interpretability}

\begin{figure}[htbp]
    \centering
    \begin{tabular}{c|c}
      \includegraphics[width=0.46\columnwidth]{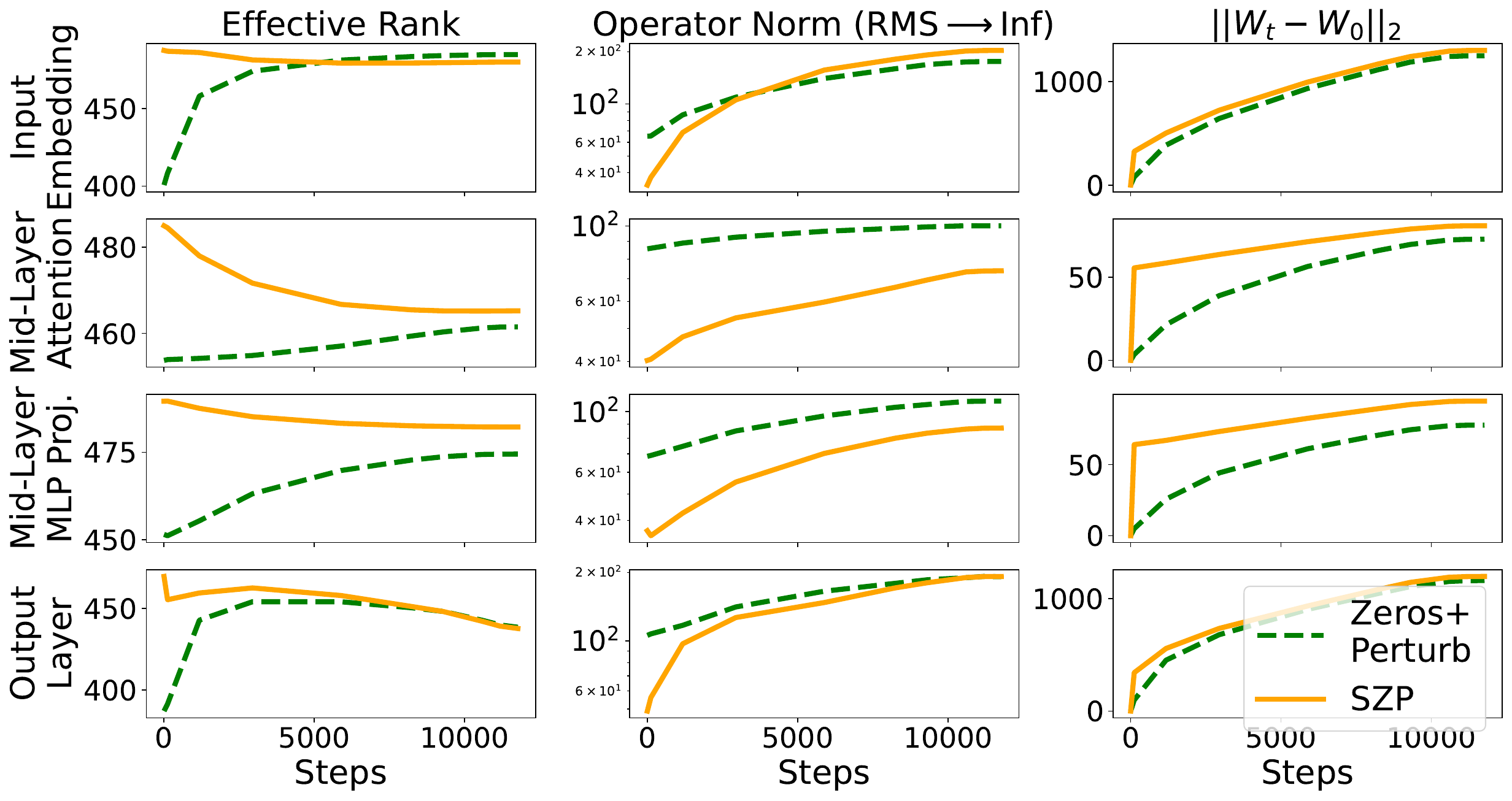} & \includegraphics[width=0.46\columnwidth]{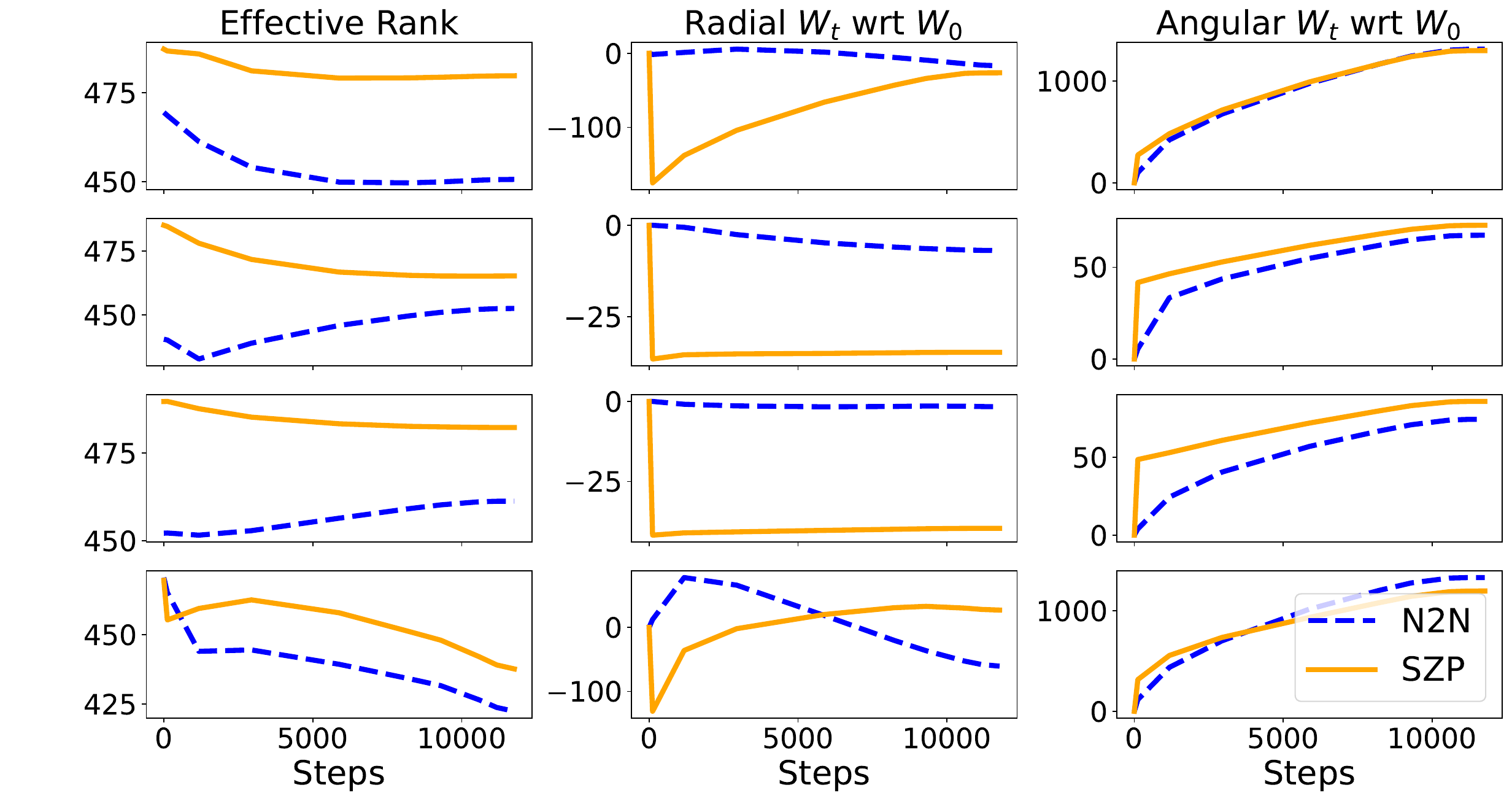} \\
    \end{tabular}
    \caption{
    Looking inside \ws{} for $14M\rightarrow{}77M$: Plotting metric traces over steps; \textit{Top row}: Input Embedding; \textit{Middle rows}: The middle layer's attention and MLP matrices; \textit{Bottom row}: Output Un-embedding; 
    (\textit{left}) Comparing \texttt{Zeros+Perturb} with \textit{Shrink+Zeros+Perturb}(\ours{}), showing the \textit{Effective Rank}, \textit{L1 Norm}, and the norm of the difference of the weights at step $t$ to the grown weight at initialization;
    (\textit{right}) Comparing \nton{} with \ours{}, showing the \textit{Effective Rank}, the norm of the weight at step $t$ projected along the unit vector at initialization, and the norm of the weight at step $t$ orthogonal to the unit vector at initialization.
    Refer to~\Cref{sec:empirical-setup} for experiment details.
    }
    \label{fig:post-hoc-inter-main}
\end{figure}

In~\Cref{fig:post-hoc-inter-main} (left), we trace weight statistics for \ours{} vs.\ \texttt{Zeros+Perturb} (\ours{} without shrinking) across the input-output layers and the middle layer's attention and MLP matrices.
The CL literature highlights shrinking's role in \textit{preserving} the learned function via weight-norm clamping, which we observe here and correlate with improved plasticity~\citep{lyle-icml23a,lyle-blog25a} and loss reduction (\Cref{fig:basic-baselines}).
As a more informative proxy than $L_1$ or $L_2$, we track the $RMS \to L_{\infty}$ operator norm — the maximum row-wise $L_2$ norm scaled by the square root of the input width — which is crucial for transferring learning dynamics across scales~\citep{filatov-arxiv25a,pethick-icml25a}.
Shrinking also improves utilization of new neurons, evidenced by the larger effective rank of the attention layer, and enables a larger shift from initialization ($||W_t - W_0||_2$), letting the grown model move further from its base weights.
This matters because zero-padding or cloning alone may not break sufficient symmetry between old and new neurons~\citep{yu-arxiv26a}, leaving the grown model too close to the base in function space, which manifests as loss saturation in~\Cref{fig:basic-baselines}.

\subsection{Perturbation Scale Ablation}
\label{app:perturbation-ablation}

Perturbation is used to activate the newly added neurons after zero-padding.
To study its sensitivity, we sweep the perturbation scale $\sigma_{\mathrm{perturb}}$ on a $32\mathrm{M}\rightarrow286\mathrm{M}$ transfer, keeping the remaining \ours{} settings fixed.
The default setting uses
\[
\sigma_{\mathrm{perturb}} = \frac{1}{\sqrt{\mathrm{width}}}.
\]

\begin{figure}[hbtp]
    \centering
    \includegraphics[width=0.45\linewidth]{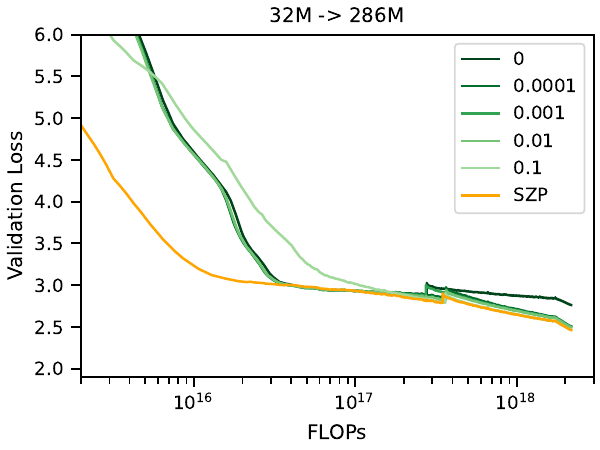}
        \caption{
    Effect of perturbation scale $\sigma_{\mathrm{perturb}}$ on a $32\mathrm{M}\rightarrow286\mathrm{M}$ transfer.
    Validation-loss trajectories are shown as a function of training FLOPs for different fixed perturbation scales.
    The \ours{} default, $\sigma_{\mathrm{perturb}}=1/\sqrt{\mathrm{width}}$, is highlighted in orange.
    Nonzero perturbations substantially improve over pure zero-padding, and the default reaches the low-loss regime quickly while matching the best final validation loss in the sweep.
    }
    \label{fig:perturbation-ablation}
\end{figure}

\Cref{tab:perturbation-ablation} reports the final validation loss for each perturbation scale, while \cref{fig:perturbation-ablation} shows the corresponding learning curves.
Without perturbation, performance is substantially worse, suggesting that pure zero-padding does not effectively activate the added capacity.
Nonzero perturbations improve both convergence and final loss, with the default $\sigma_{\mathrm{perturb}}=1/\sqrt{\mathrm{width}}$ matching the best final performance in this sweep.
The learning curves further show that the default reaches the low-loss region earlier than most fixed perturbation scales, which supports using it as a simple scale-aware choice rather than tuning $\sigma_{\mathrm{perturb}}$ separately for each transfer.

\begin{table}[t]
\centering
\caption{
Perturbation-scale ablation for \ours{} on a $32\mathrm{M}\rightarrow286\mathrm{M}$ transfer.
The table reports final validation loss; the corresponding learning curves are shown in \cref{fig:perturbation-ablation}.
}
\label{tab:perturbation-ablation}
\begin{tabular}{lc}
\toprule
\textbf{Hyperparameter} & \textbf{Final validation loss} \\
\midrule
\multicolumn{2}{l}{\textbf{Perturbation scale $\sigma_{\mathrm{perturb}}$}} \\
\midrule
$0$        & $2.76$ \\
$10^{-4}$  & $2.50$ \\
$10^{-3}$  & $2.50$ \\
$10^{-2}$  & $2.49$ \\
$10^{-1}$  & $2.46$ \\
\textbf{\ours{}} $(1/\sqrt{\mathrm{width}})$ & $\mathbf{2.46}$ \\
\bottomrule
\end{tabular}
\end{table}

\subsection{Shrinkage Ablation}
\label{app:shrinking-ablation}

Shrinkage controls the strength of the copied base-model path in \ours{}.
In the initialization 
$\Theta_0=\lambda_{\mathrm{shrink}}P(\theta^\star)+E$, smaller values of
$\lambda_{\mathrm{shrink}}$ reduce the contribution of the inherited narrow solution, while the perturbation $E$ activates the newly added coordinates.
As discussed in \Cref{app:szp-theory}, this makes it easier for the new neurons to compete with the copied subnetwork after growth.

This role of shrinkage is consistent with recent work on plasticity in warm-started networks.
DASH~\citep{shin-icml24a} also uses shrinkage to mitigate plasticity loss, but does so in a direction-aware manner: weights associated with useful learned features are preserved more strongly, while components attributed to memorized noise are shrunk more aggressively.
In contrast, \ours{} intentionally uses a single global shrinkage factor, making it architecture-agnostic and easy to combine with arbitrary growth operators.
Relatedly, weight clipping has been proposed as a simple way to control weight magnitudes and improve plasticity in continual and reinforcement learning settings~\citep{elsayed-rlc24}. 
Our use of shrinkage can be viewed as a one-shot initialization-time analogue of this broader norm-control principle.

\begin{figure}[hbtp]
    \centering
    \includegraphics[width=0.75\linewidth]{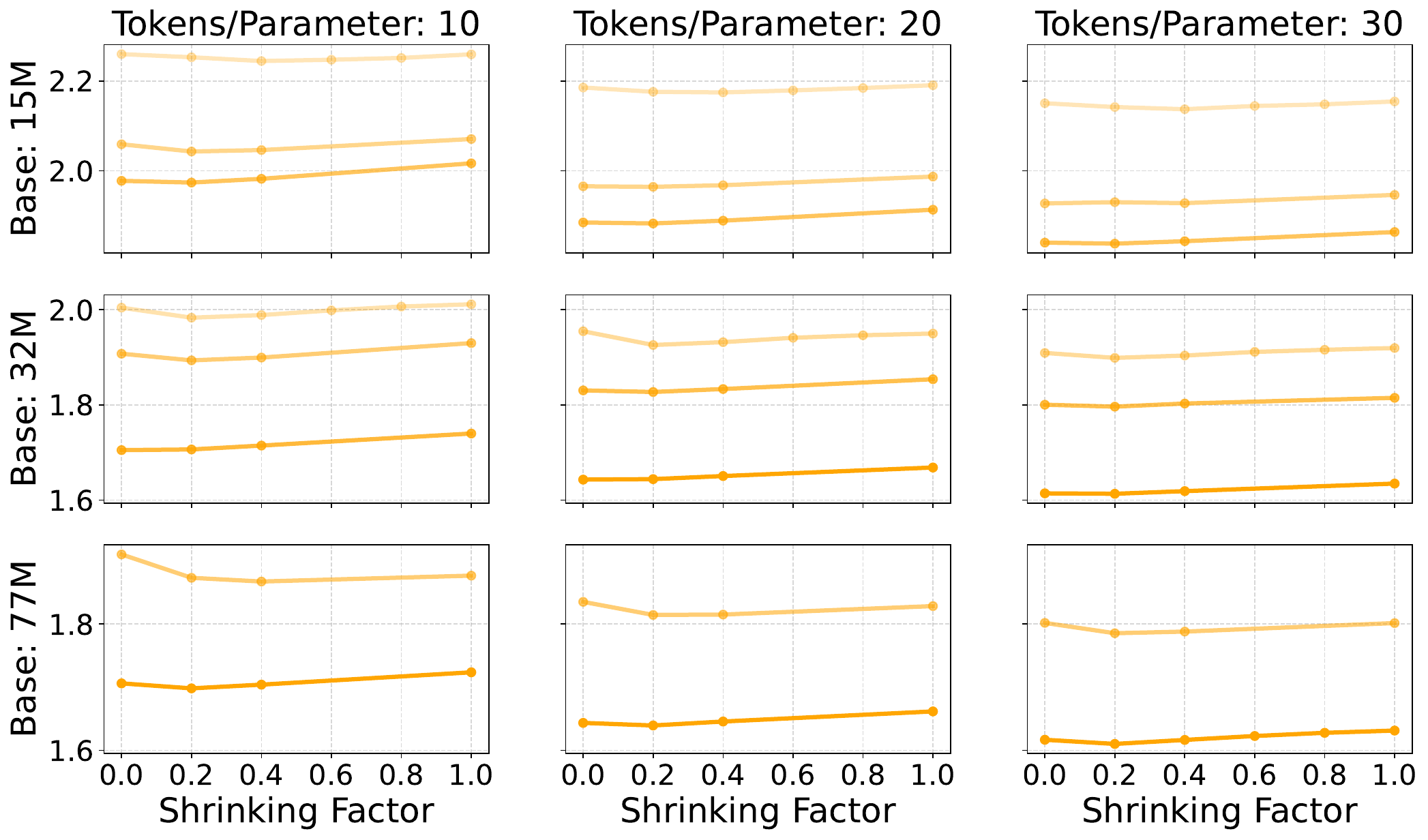}
    \caption{
    Ablation for shrinking factor ($\shrinkhp$) for \ours{}. 
    The rows represent a base checkpoint. 
    Each line in the plot represents a target model scale warmstarted from the given checkpoint for the row.
    The darker the line, the larger the target size (larger growth factor).}
    \label{fig:shrinking-ablation}
\end{figure}

We ablate $\lambda_{\mathrm{shrink}}$ while keeping the remaining \ours{} settings fixed.
\Cref{fig:shrinking-ablation} reports final validation loss across three base checkpoints, three token budgets, and multiple target scales.
Across settings, the best values are concentrated in the low-to-moderate shrinkage regime, typically around $\lambda_{\mathrm{shrink}}\in\{0.2,0.4\}$.
Setting $\lambda_{\mathrm{shrink}}$ too close to $1$ corresponds to little or no shrinkage, leaving the grown model more dominated by the inherited base-model path.
Conversely, setting it too close to $0$ removes too much of the pretrained signal and makes the initialization closer to perturbation-only growth.
The broad minimum around $0.2$--$0.4$ suggests that shrinkage is useful, but does not require precise tuning.

We use $\lambda_{\mathrm{shrink}}=0.4$ in the main experiments as a stable default in this low-loss region.
This choice is also consistent with the post-hoc analysis in \Cref{fig:post-hoc-inter-main}: compared with perturbation alone, adding shrinkage improves effective rank and permits larger movement from initialization, indicating that the grown model better uses its added capacity rather than remaining anchored to the base model.

\section{Language Modeling Task Details}
\label{app:llm-details}

\subsection{Model Scales}\label{app:model_scales}

\begin{table}[t]
    \centering
    \caption{
    Language-model scales used in our experiments.
    All models have $8$ layers and a sequence length $1024$. We scale the width while keeping the attention head size fixed at $64$.
    Parameter counts are rounded to the nearest million.
    }
    \label{tab:model_scaling}
    \small
    \setlength{\tabcolsep}{8pt}
    \renewcommand{\arraystretch}{1.05}
    \begin{tabular}{cccc}
    \toprule
    \textbf{$d_{\mathrm{model}}$} & \textbf{$n_{\mathrm{head}}$} & \textbf{Head size} & \textbf{Params (M)} \\
    \midrule
    $128$  & $2$  & $64$ & $14$  \\
    $256$  & $4$  & $64$ & $32$  \\
    $512$  & $8$  & $64$ & $77$  \\
    $768$  & $12$ & $64$ & $134$ \\
    $1280$ & $20$ & $64$ & $286$ \\
    $2048$ & $32$ & $64$ & $610$ \\
    $3072$ & $48$ & $64$ & $1200$ \\
    \bottomrule
    \end{tabular}
\end{table}

Table~\ref{tab:model_scaling} summarizes the language-model scales used across our experiments.

In this work, we focus only on width-scaling, specifically the setting where the \textit{head size}, i.e., the output dimension of each attention head, remains fixed across scales.
Equivalently, this form of width-scaling adds \textit{more heads} of the same size to each layer.
This approach is standard in the literature, especially when studying feature learning under width-scaling~\citep{everett-icml24a}.
We note that this form of width-scaling provides the fairest setup for comparing \ws{} methods.
The \textit{query}, \textit{key}, and \textit{value} weights are grouped per head within a tensor.

For a simple mental illustration, a $1$-head network with width $1$ will have an attention tensor with $3$ dimensions, one each for \textit{query}, \textit{key}, and \textit{value}, assuming no weight sharing.
When grown to width $2$ by scaling the head dimension, the flattened attention tensor has $6$ dimensions, $2$ each for \textit{query}, \textit{key}, and \textit{value}.
This introduces a subtle layout inefficiency: under index-wise growth, the second dimension of the \textit{query} block in the larger model corresponds to the \textit{key} dimension of the smaller model.
This correspondence is maintained cleanly when width-scaling is implemented by increasing only the number of heads while keeping the head size fixed.

\subsection{Optimization Details}\label{app:opt}

\paragraph{Optimizer.} We optimize the cross--entropy loss with AdamW~\citep{loshchilov-iclr19a}. Unless otherwise stated, we keep the original moments $(\beta_1,\beta_2)=(0.9,0.95)$ and numerical stabilizer $\epsilon=10^{-8}$. Gradient global norm clipping is set to~$1.0$.

\paragraph{Adam $\epsilon$.} 
We set it to the PyTorch default of $10^{-8}$ that also follows the suggestion from~\citet{everett-icml24a} for sub-billion parameter models.

\paragraph{Weight Decay.} All experiments use \emph{zero} weight decay. This isolates the effect of warmstarting from explicit $\ell_2$~regularization, avoiding confounding interactions between the two mechanisms.

\paragraph{Warmup--Stable--Decay (WSD) Schedule.} The learning rate follows a trapezoidal profile - let $T$ be the total number of optimizer steps. For $\eta_{\max}$ the peak learning rate, warm-up fraction $w=0.01$ and decay fraction $d=0.20$, the instantaneous rate $\eta_t$ is
\[
  \eta_t =
  \begin{cases}
    \eta_{\max}\,\tfrac{t}{wT}, & 0 \le t < wT, \\\\
    \eta_{\max}, & wT \le t < (1-d)T, \\\\
    \eta_{\max}\,\tfrac{T-t}{dT}, & (1-d)T \le t \le T.
  \end{cases}
\]
Warm-up ensures numerical stability in the early regime~\citep{kosson2024analyzing}, while the constant plateau facilitates long, steady optimization at a single effective step size. The linear cool-down to zero replaces the abrupt drop used by step schedules and was found to improve final performance by~\citet{hagele-neurips24a}.

\paragraph{Training Budget.}
Unless otherwise stated, language-model runs use budgets of $10$, $20$, or $30$ tokens per model
parameter, with $20$ \tkpm{} corresponding to the Chinchilla compute-optimal reference
budget~\citep{hoffmann-neurips22a}. For a given model scale, the number of optimizer steps is
computed from the total token budget and the effective batch size in \Cref{tab:lm-hpo-transfer}.

\subsection{Grid Search Results for Optimal Hyperparameters}
\label{app:grid_search}

For the language-model experiments, we tune the peak learning rate $\eta_{\max}$ and effective batch size at small base scales, then transfer the selected configuration to larger target scales using the $\mup{}$ rules. 
This section only reports the base-scale choices used for transfer. 
The model sizes are listed in \Cref{tab:model_scaling}, and the shared optimizer and schedule settings are given in \Cref{app:opt}.

\begin{table}[t]
\centering
\caption{Base-scale language-model hyperparameters selected for $\mup{}$ transfer.}
\label{tab:lm-hpo-transfer}
\small
\setlength{\tabcolsep}{8pt}
\renewcommand{\arraystretch}{1.08}
\begin{tabular}{lcc}
\toprule
\textbf{Params (M)} & \textbf{Selected $\eta_{\max}$} & \textbf{Effective batch size} \\
\midrule
$14$ & $3\times 10^{-3}$   & $64$ \\
$32$ & $5\times 10^{-3}$   & $128$ \\
$77$ & $2.5\times 10^{-3}$ & $256$ \\
\bottomrule
\end{tabular}
\end{table}

For target-scale runs, per-device micro-batch sizes are chosen as the largest values that fit in memory on the available L40S GPUs, and gradient accumulation is used to realize the transferred effective batch sizes. The same base-scale selections are reused across the $10$, $20$, and $30$ \tkpm{} budgets, with the number of optimizer steps adjusted according to the budget.

\section{Compute Resources}\label{app:compute}

We report approximate compute for the experiments used in the paper in
\Cref{tab:cpu-compute-by-scale,tab:cpu-compute-additional-tkpm,tab:gpu-compute-by-scale}.
The synthetic MLP experiments were run on CPU nodes with dual-socket AMD EPYC 9655 processors, providing 192 physical cores in total.
The language-model experiments were run primarily on NVIDIA L40S GPUs with 48GB VRAM using CUDA~12.7, with the $1.2$B runs executed on H200 GPUs.

For the synthetic MLP experiments, \Cref{tab:cpu-compute-by-scale} reports the $20$ \tkpm{} runs, which include the base grid, main target-width grid, and transfer runs described in Appendix~\ref{app:regression_hpspaces}.
Additional $30$ and $50$ \tkpm{} MLP runs are transfer-only and are used for the IsoFLOP fits. For each budget, this gives $378$ runs. 
These additional runs are summarized in \Cref{tab:cpu-compute-additional-tkpm}.
Together, the reported synthetic MLP experiments account for approximately $10{,}143$ CPU-hours, or $423$ CPU-days.

For the reported language-model experiments, we report GPU-hour ranges based on the number of completed runs, hardware allocation, and typical wall-clock time per target scale.
Including an allowance for debug and re-runs, this corresponds to roughly $39{,}000$--$50{,}000$ GPU-hours.

These estimates aggregate the completed runs used in the reported figures and exclude exploratory, failed, or auxiliary runs not used in the paper.
The associated emissions footprint is expected to be lower than implied by raw compute alone, as the hosting facility is powered primarily by renewable electricity.

\begin{table}[t]
\centering
\caption{Approximate GPU compute used by target model scale for the reported language-model experiments. Unless otherwise noted, runs use NVIDIA L40S GPUs; the 1.2B runs use H200 GPUs.}
\label{tab:gpu-compute-by-scale}
\small
\setlength{\tabcolsep}{5pt}
\renewcommand{\arraystretch}{1.05}
\begin{tabular}{lccc}
\toprule
\textbf{Target scale (M)} & \textbf{\# runs} & \textbf{GPU-days / run} & \textbf{Total GPU-hours} \\
\midrule
$32$        & $25$ & $1.5-2.0$  & $900-1200$ \\
$77$        & $40$ & $2.0-3.0$  & $1920-2880$ \\
$134$       & $49$ & $4.0-6.5$  & $4700-7650$ \\
$286$       & $53$ & $12-16$    & $15300-20400$ \\
$610$       & $13$ & $32$        & $10000$ \\
$1200$ (H200) & $2$ & $56-64$    & $2700-3100$ \\
\midrule
\textbf{Subtotal} & \textbf{$182$} & -- & $35600-45200$ \\
\textbf{$+\text{debug/re-runs}(\sim{}10\%)$} & -- & -- & \textbf{$\sim{}450.0$} \\
\midrule
\textbf{Total} & -- & -- & $39000-50000$ \\
\bottomrule
\end{tabular}
\end{table}

\begin{table}[t]
\centering
\caption{Approximate CPU compute used by target width for the reported $20$ \tkpm{} width-scaling synthetic MLP experiments. This includes the base grid, target-width grid, and transfer runs.}
\label{tab:cpu-compute-by-scale}
\small
\setlength{\tabcolsep}{6pt}
\renewcommand{\arraystretch}{1.05}
\begin{tabular}{lcccc}
\toprule
\textbf{Width} & \textbf{\# base widths} & \textbf{\# runs} & \textbf{CPU-hours / run} & \textbf{Total CPU-hours} \\
\midrule
$48$  & -- & $720$ & $0.17$ & $120.0$ \\
$68$  & $1$  & $234$ & $0.27$ & $62.4$ \\
$96$  & $2$  & $252$ & $0.48$ & $119.7$ \\
$136$ & $3$  & $270$ & $0.92$ & $247.5$ \\
$192$ & $4$  & $288$ & $1.81$ & $520.0$ \\
$272$ & $5$  & $306$ & $4.31$ & $1317.5$ \\
$384$ & $6$  & $108$ & $10.00$ & $1080.0$ \\
\midrule
\textbf{Total} & -- & \textbf{$2178$} & -- & \textbf{$3467.1$} \\
\bottomrule
\end{tabular}
\end{table}

\begin{table}[t]
\centering
\caption{Additional CPU compute for transfer-only synthetic MLP runs used in the IsoFLOP fits. 
}
\label{tab:cpu-compute-additional-tkpm}
\small
\setlength{\tabcolsep}{8pt}
\renewcommand{\arraystretch}{1.05}
\begin{tabular}{lcc}
\toprule
\textbf{Budget} & \textbf{\# runs} & \textbf{Total CPU-hours} \\
\midrule
$30$ \tkpm{} & $378$ & $2503.4$  \\
$50$ \tkpm{} & $378$ & $4172.3$ \\
\midrule
\textbf{Total} & \textbf{$756$} & \textbf{$6675.6$}  \\
\bottomrule
\end{tabular}
\end{table}

\section{Additional Empirical Results}
\label{app:more-results}

\subsection{LLaMa Architecture Results}
\label{app:llama-metrics}

\begin{figure}[hbtp]
    \centering
    \includegraphics[width=0.75\linewidth]{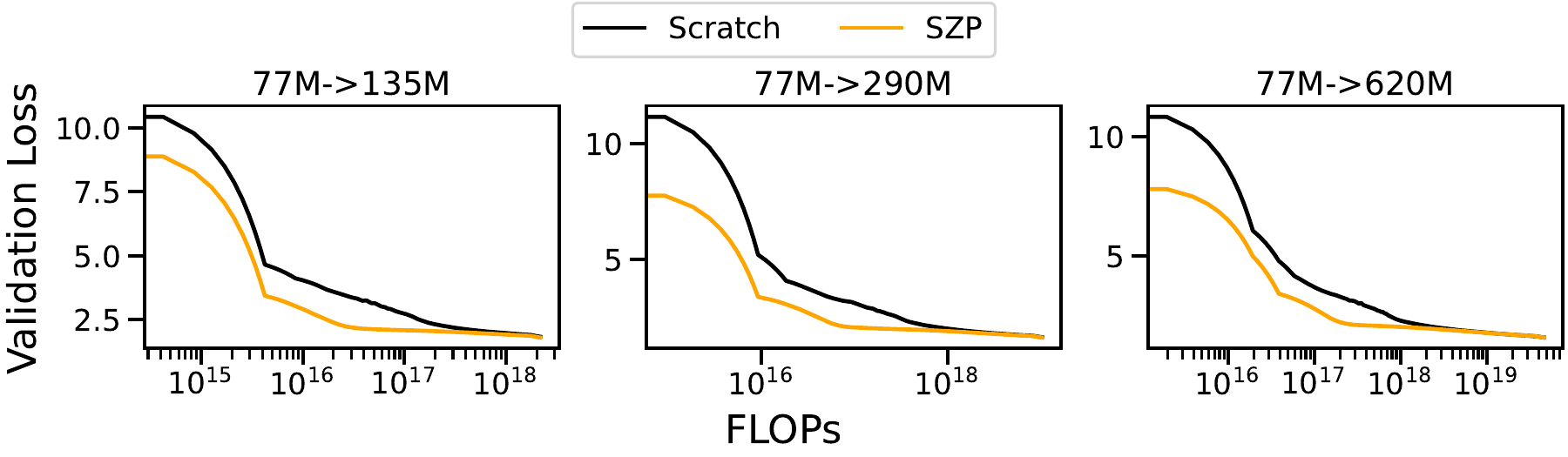}
    \caption{
    Comparing \ours{} with \scratch{} on a LLaMA-style decoder architecture~\citep{touvron2023llama}, trained for $20$ \tkpm{}. 
    Both methods use the same base-scale hyperparameter selection and transfer protocol; they differ only in initialization. 
    \ours{} warmstarts the target model from the corresponding base checkpoint, while \scratch{} trains the target model from a random initialization.}
    \label{fig:llama}
\end{figure}

To test whether the behavior of \ours{} is specific to the GPT-2 style architecture used in the main language-model experiments, we repeat the comparison on a LLaMA-style decoder architecture \citep{touvron2023llama}. We keep the same training budget of $20$ \tkpm{} and use the same
base-scale hyperparameter selection protocol for both \scratch{} and \ours{}. 
The only difference between the two runs is the initialization: \scratch{} is trained from a randomly initialized target model, while \ours{} initializes the target model by warmstarting from the corresponding base
checkpoint. 
As shown in \Cref{fig:llama}, \ours{} consistently improves early convergence across all tested target sizes, suggesting that the gains are not tied to a particular decoder implementation.

\subsection{Downstream Task Results}
\label{app:downstream-metrics}

To complement the validation-loss comparison in \Cref{sec:empirical-practice}, we evaluate the same $1.2$B-parameter \scratch{} and \ours{} checkpoints on a compact suite of downstream multiple-choice benchmarks using \textsc{LightEval}~\citep{lighteval}.
These evaluations are intended as a sanity check that the gains from warmstarting are not limited to language-model validation loss, but also translate to standard commonsense, science, and general knowledge tasks.
We report accuracy on ARC-Challenge~\citep{clark2018think}, HellaSwag~\citep{zellers2019hellaswag}, MMLU~\citep{hendrycks2021measuring}, OpenBookQA~\citep{mihaylov2018can}, PIQA~\citep{bisk2020piqa}, and WinoGrande~\citep{sakaguchi2020winogrande} in \Cref{tab:downstream-results}.
For the subset overlapping with Hugging Face Open LLM Leaderboard v1, we follow the corresponding fixed few-shot settings: ARC-Challenge $25$-shot, HellaSwag $10$-shot, MMLU $5$-shot, and WinoGrande $5$-shot~\citep{hf-open-llm-leaderboard-v1,lighteval-leaderboard-tasks}.
We additionally include OpenBookQA and PIQA as standard \textsc{LightEval} multiple-choice tasks in the $0$-shot setting.
Across all six benchmarks, \ours{} improves over the corresponding \scratch{} baseline, suggesting that the optimization advantage observed in the pretraining curves carries over to downstream behavior.

Since these results are based on a single checkpoint per method and do not include multiple training seeds or statistical significance testing, we interpret them as a sanity check rather than a definitive downstream-performance claim.

\begin{table}[htbp]
\centering
\caption{Downstream multiple-choice accuracy for the $1.2$B-parameter checkpoints. Higher is better.}
\label{tab:downstream-results}
\small
\setlength{\tabcolsep}{4pt}
\renewcommand{\arraystretch}{1.05}
\begin{tabular}{lcccccc}
\toprule
\textbf{Method} & \textbf{ARC-C} & \textbf{HSwag} & \textbf{MMLU} & \textbf{OBQA} & \textbf{PIQA} & \textbf{WinoG.} \\
\midrule
\scratch{} & $17.7$ & $27.4$ & $26.0$ & $13.4$ & $60.2$ & $51.0$ \\
\ours{}    & $\textbf{20.1}$ & $\textbf{29.4}$ & $\textbf{26.4}$ & $\textbf{15.8}$ & $\textbf{64.6}$ & $\textbf{53.0}$ \\
\bottomrule
\end{tabular}
\end{table}

\section{$\nton$}
\label{app:net2net}

\nton{}~\citep{chen-iclr16a} is a seminal function-preserving growth method: it expands a trained \emph{base} network into a larger \emph{target} network by duplicating/splitting hidden units (width) and/or inserting identity layers (depth), such that the target initially computes (approximately) the same function as the base.
In recent standardized evaluations of model growth for LLM pre-training, a \nton{}-style \emph{direct} width-expansion operator is used as the canonical function-preserving width-growth baseline (often referred to as a ``direct'' growth operator)~\citep{du-neurips24a}.
We therefore adopt \nton{} as our primary width warmstarting baseline.

\subsection{Why Net2Net?}
\label{app:net2net-choice}

We use \nton{}~\citep{chen-iclr16a} because it (1) targets \emph{width expansion}, which is the focus of this work, (2) is (approximately) function-preserving at initialization, and (3) requires no changes to the training objective or additional optimization stages.
This makes it a clean and widely used baseline for isolating the effect of the warmstarted initialization.
Moreover, in the large-scale growth benchmark of \citet{du-neurips24a}, the Net2Net-style \emph{direct} widening operator is the strongest width-expansion baseline among the compared growth operators, motivating its use as our primary width warmstarting baseline.

\paragraph{Why not other Methods?}
We exclude several alternative strategies by design. Learned growth and distillation-based methods (for example, linear transformations~\citep{wang-iclr23a}, progressive module training~\citep{gong-icml19a}, or teacher--student frameworks~\citep{qin-arxiv21a}) introduce additional stages (e.g., auxiliary alignment losses or intermediate supervision), which makes it harder to attribute gains specifically to initialization quality.
Importantly, our method could be composed with such techniques; however, such extensions are outside the scope of this work.

\subsection{Net2Net Width Expansion}
\label{app:net2net-formalism}

\begin{figure}[t]
    \centering
    \includegraphics[width=0.78\linewidth]{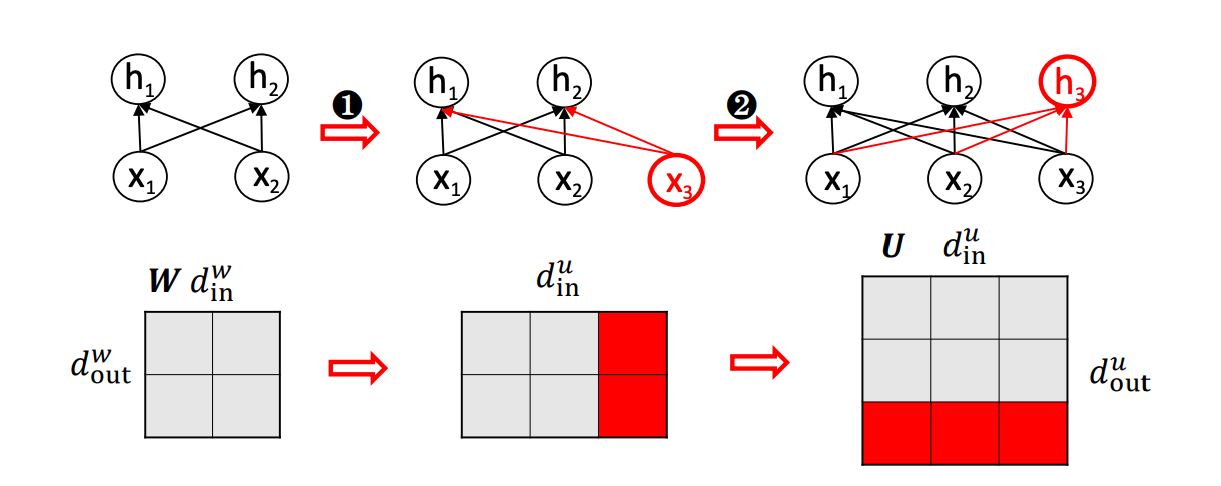}
    \caption{\textbf{Net2Net-style matrix expansion.} A width expansion can be implemented by duplicating columns (fan-in growth) and rows (fan-out growth), corresponding to adding new hidden units (highlighted). We use this figure (adapted from transformer growth descriptions building on \nton{} and \btob{}) to illustrate the operation. Sourced from~\citep{chen-acl21a}. }
    \label{fig:net2net-matrix-expansion}
\end{figure}

\paragraph{Intuition.}
Width growth in \nton{} can be viewed as \emph{splitting} a hidden unit into several identical copies, then rescaling the outgoing weights so that the combined contribution of the copies matches the original unit.
\Cref{fig:net2net-matrix-expansion} illustrates this matrix-level perspective: we first expand the input dimension (duplicate columns), then expand the output dimension (duplicate rows).

\paragraph{Neuron-splitting Formalism (single hidden layer).}
Consider a base MLP block with hidden width $n$,
\[
h = \phi(W_{\text{in}} x), \qquad y = W_{\text{out}} h,
\]
where $W_{\text{in}} \in \mathbb{R}^{n \times d}$ and $W_{\text{out}} \in \mathbb{R}^{d_{\text{out}} \times n}$.
To grow the hidden width from $n$ to $q$, \nton{} defines a mapping $\pi \in \{1,\dots,n\}^{q}$ that assigns each new unit $j$ to a source unit $\pi(j)$.
Let $c_i = |\{j : \pi(j)=i\}|$ be the copy count of base unit $i$.
Then a function-preserving widening is achieved by,
\begin{align}
(W_{\text{in}}')_{j,:} &= (W_{\text{in}})_{\pi(j),:}, \\
(W_{\text{out}}')_{:,j} &= \tfrac{1}{c_{\pi(j)}} (W_{\text{out}})_{:,\pi(j)}.
\end{align}
Immediately after expansion, all copies produce identical activations, and the rescaled outgoing weights ensure that the sum of their contributions equals the original unit, so $W'_{\text{out}}\phi(W'_{\text{in}}x)=W_{\text{out}}\phi(W_{\text{in}}x)$.

\paragraph{Matrix Form (general linear layers).}
The same idea can be written compactly at the matrix level.
For a linear map $W \in \mathbb{R}^{d_{\text{out}}\times d_{\text{in}}}$ expanded to
$W' \in \mathbb{R}^{D_{\text{out}}\times D_{\text{in}}}$, Net2Net-style widening can be expressed as
\begin{equation}
W' = L \, W \, R,
\label{eq:net2net-lwr}
\end{equation}
where $R$ duplicates (copies) input dimensions, i.e., columns of $W$, to reach $D_{\text{in}}$,
and $L$ duplicates output dimensions, i.e., rows of $W$, to reach $D_{\text{out}}$.
To preserve the function, $L$ and/or $R$ additionally include a $1/c$ rescaling for duplicated units, where $c$ is the copy count of the corresponding source unit.
This is the same ``copy-and-rescale'' operator view commonly used when describing width growth in transformer blocks~\citep{du-neurips24a}.

\paragraph{Noise for Symmetry Breaking.}
Because duplicated units start identically, we optionally add small Gaussian noise to the newly created parameters (or to a subset of duplicated rows/columns) to break symmetry, following standard \nton{} practice~\citep{chen-iclr16a}.
In our experiments this noise is small enough to keep the initialization close to function-preserving while avoiding perfectly identical copies.

\subsection{Implementation in our Experiments}
\label{app:net2net-impl}

\paragraph{MLPs.}
In the synthetic MLP experiments, we apply the widening rule above to each hidden layer: we duplicate hidden units according to $\pi$ and rescale the corresponding outgoing weights by the copy counts, optionally injecting small Gaussian noise for symmetry breaking.

\paragraph{Transformers.}
In the transformer language-model experiments, we apply the same function-preserving principle when expanding width.
Concretely, we grow matrices whose inner dimensions scale with model width (e.g., token embeddings and the projection matrices in attention/MLP blocks) via the same ``copy columns / copy rows + rescale'' recipe in \Cref{eq:net2net-lwr}.
For multi-head attention, because we keep the head size fixed and grow width by adding heads, copying is performed at the level of complete heads: duplicated $W_q$, $W_k$, and $W_v$ rows are left unscaled, while the copy-count rescaling is applied only to the corresponding $W_o$ columns.
This preserves the duplicated heads' softmax computation while averaging their combined contribution through the output projection.
This yields an initialization that matches the base model computation as closely as possible at the moment of expansion, while remaining a drop-in replacement that does not modify the objective or training loop.

All comparisons use identical token budgets and optimization settings to isolate the effect of initialization.

\subsection{Scope and Follow-up Work}
\label{app:net2net-width-only}

While our study focuses on width-only growth, \nton{}~\citep{chen-iclr16a} also proposes depth-expanding transformations (e.g., inserting identity layers).
In transformers, \btob{}~\citep{chen-acl21a} extends \nton{}-style ideas to BERT, and combines them with layer stacking to grow depth.
More broadly, depth expansion can be paired with transformer-specific stacking strategies, such as those explored in \emph{Stacking Your Transformers}~\citep{du-neurips24a}.

Our empirical study isolates \emph{width scaling} (following~\citet{everett-icml24a}) because our emphasis on hyperparameter transfer and scaling behavior requires extensive sweeps across widths, training horizons, and transfer settings.
Nevertheless, extending our warmstarting analysis to depth is a natural next step, and should be compatible with depth-stacking mechanisms and depth-aware scaling rules (e.g., depth $\mup$ variants and recipe-based transfers such as completeP~\citep{dey-arxiv25a}).
Ultimately, this points toward studying warmstarting under compound scaling; we first establish controlled evidence for hyperparameter transfer under width growth~\citep{everett-icml24a,dey-arxiv25a,essentialai-arxiv25a}.

\subsection{Comparison with \hc{}}
\label{app:hypercloning}

\begin{table}[htbp]
\caption{
Warmstarting from a \emph{base}$\rightarrow$\emph{target} model size, trained for $20$ \tkpm{}, comparing \hc{} to \ours{} for up to $287$M parameters. Both start from the same base checkpoints with $\mup{}$-transferred hyperparameters and differ only in the growth operator. Validation loss at four matched FLOP levels per pair; the better value in bold. \ours{} reaches lower loss at every FLOP level, with the gap largest early in training.
}
\label{tab:hypercloning}
\centering
\small
\setlength{\tabcolsep}{4.5pt}
\begin{tabular}{rcc rcc rcc}
\toprule
\multicolumn{3}{c}{$32$M $\rightarrow$ $77$M} & \multicolumn{3}{c}{$32$M $\rightarrow$ $135$M} & \multicolumn{3}{c}{$77$M $\rightarrow$ $287$M} \\
\cmidrule(lr){1-3} \cmidrule(lr){4-6} \cmidrule(lr){7-9}
FLOPs & \hc{} & \ours{} & FLOPs & \hc{} & \ours{} & FLOPs & \hc{} & \ours{} \\
\midrule
1.5e15 & 3.78 & \textbf{3.26} & 7.6e15 & 4.08 & \textbf{3.01} & 2.1e16 & 4.18 & \textbf{3.14} \\
1.1e16 & 2.89 & \textbf{2.32} & 5.0e16 & 2.98 & \textbf{2.24} & 1.6e17 & 2.81 & \textbf{2.02} \\
9.1e16 & 2.25 & \textbf{2.19} & 3.3e17 & 2.24 & \textbf{2.08} & 1.2e18 & 2.07 & \textbf{1.87} \\
7.1e17 & 1.98 & \textbf{1.93} & 2.2e18 & 1.93 & \textbf{1.83} & 9.1e18 & 1.69 & \textbf{1.67} \\
\bottomrule
\end{tabular}
\end{table}

\nton{} is the canonical function-preserving width operator but predates Transformers. 
To check that our conclusions are not tied to it, we compare \ours{} against \hc{}~\citep{samragh-arxiv24a}, a recent function-preserving initialization for dense Transformers that tiles the base weights to fill the target width and rescales them so that the grown model reproduces the base model's outputs at initialization. 
Both methods start from the same base checkpoints, train for $20$ \tkpm{}, and use the same $\mup{}$-transferred hyperparameters (Section~\ref{sec:exp-lm-setup}); the runs differ only in the growth operator.

Table~\ref{tab:hypercloning} reports validation loss at four matched FLOP levels for three base$\rightarrow$target pairs. 
\ours{} reaches lower loss at every FLOP level. 
The gap is largest early in training and narrows as training proceeds, as in the \nton{} comparison in Figure~\ref{fig:n2n-snp-main}. 
We attribute the gap to the same symmetry argument as for \nton{} (Appendix~\ref{app:interpretability}): cloned blocks start as exact duplicates, whereas \ours{} breaks symmetry at initialization. 
Exact function preservation additionally requires the target width to be an integer multiple of the base width; for the $512\rightarrow1280$ pair ($2.5\times$) we fill the remaining $256$ columns with a half-tile of the base weights, so this run is function-preserving only \textit{approximately}.

\section{Limits of Effective Warmstarting}
\label{app:scaling-laws}
\subsection{Extended IsoFLOP Analysis}
\label{app:more_isoflops}

\paragraph{Mechanistic Interpretability of the Growth Factor.}

\begin{figure}[htbp]
    \centering
    \includegraphics[width=0.95\linewidth]{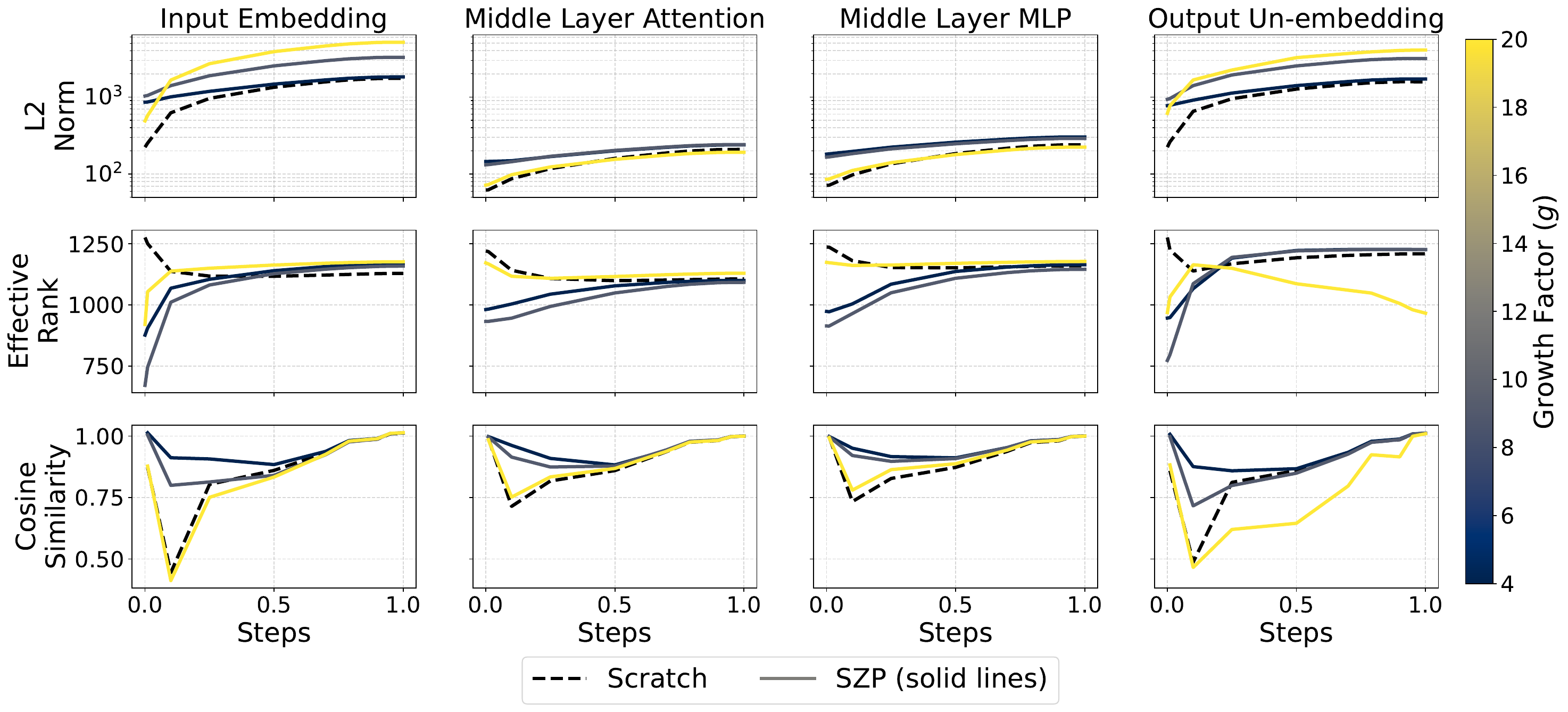}
    \caption{Mechanistic Interpretability for varying growth factors $g$ for a 286M parameter LLM trained with 20 \tkpm{},
    tracked across the input embedding, middle layer's attention and MLP matrices, and output unembedding.
    Solid lines show SZP at varying $g$ (color scale), the dashed line shows training from scratch ($g=1$). Cosine similarity is computed between consecutive checkpoints. At higher growth factors, \WS{} increasingly resembles \scratch{}.
    }
    \label{fig:interpret_over_g}
\end{figure}

\Cref{fig:interpret_over_g} looks at the trace of weight statistics given increasing growth factor.
The warmstarted models at higher $g$ increasingly resemble \scratch{} across interpretability metrics, suggesting gradual dilution of the base model's structure and explaining the degradation in performance.

\begin{figure}[htbp]
    \centering
    \includegraphics[width=\linewidth]{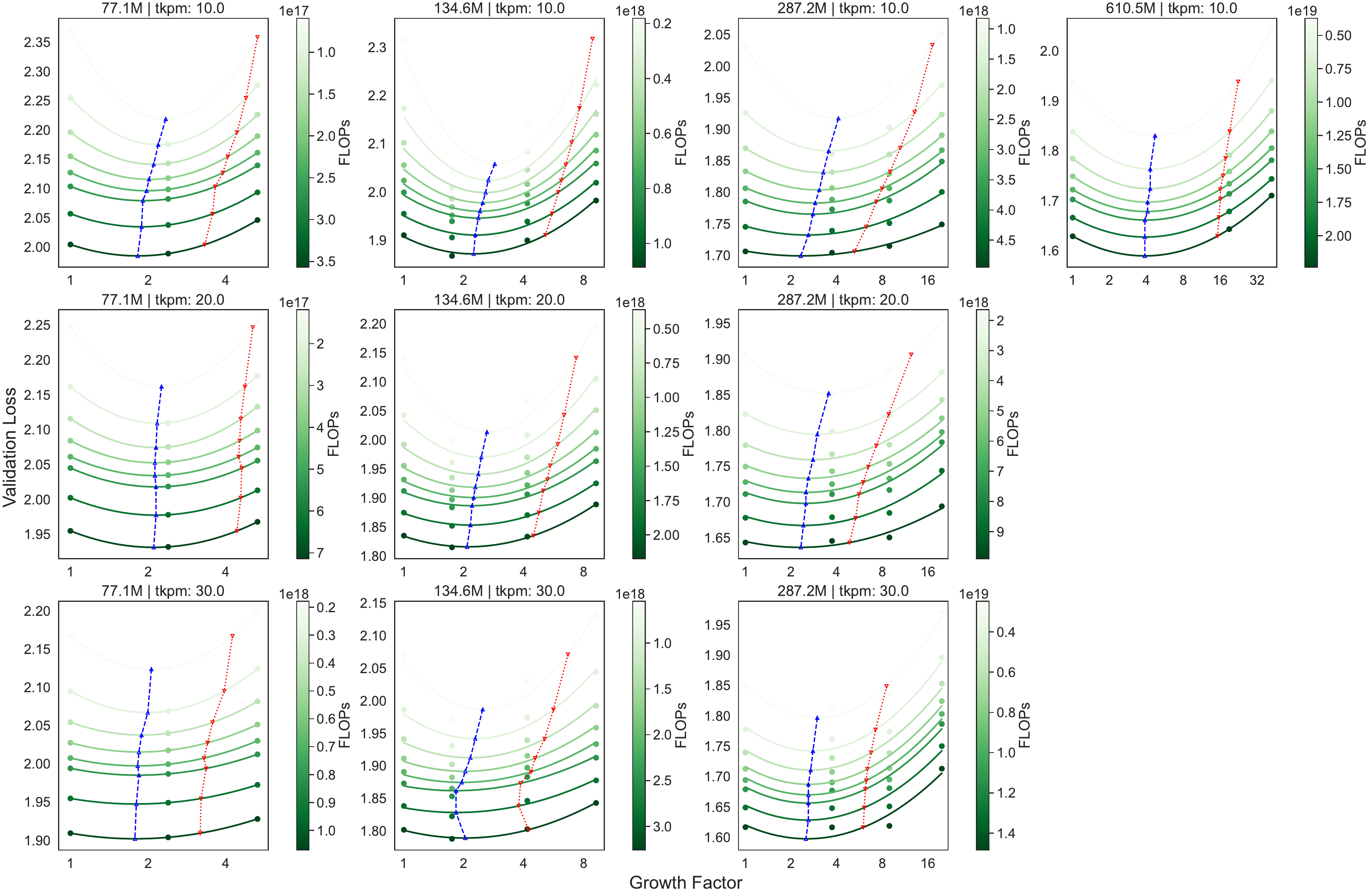}
    \caption{IsoFLOPs for \ours{} in the LLM setting. Blue and red lines denote $\gopt$ and $\gub$ respectively, as in \cref{fig:isoflops-main}.}
    \label{fig:isoflops_llm_szp_app}
\end{figure}

\begin{figure}%
    \centering
    \includegraphics[width=0.8\linewidth]{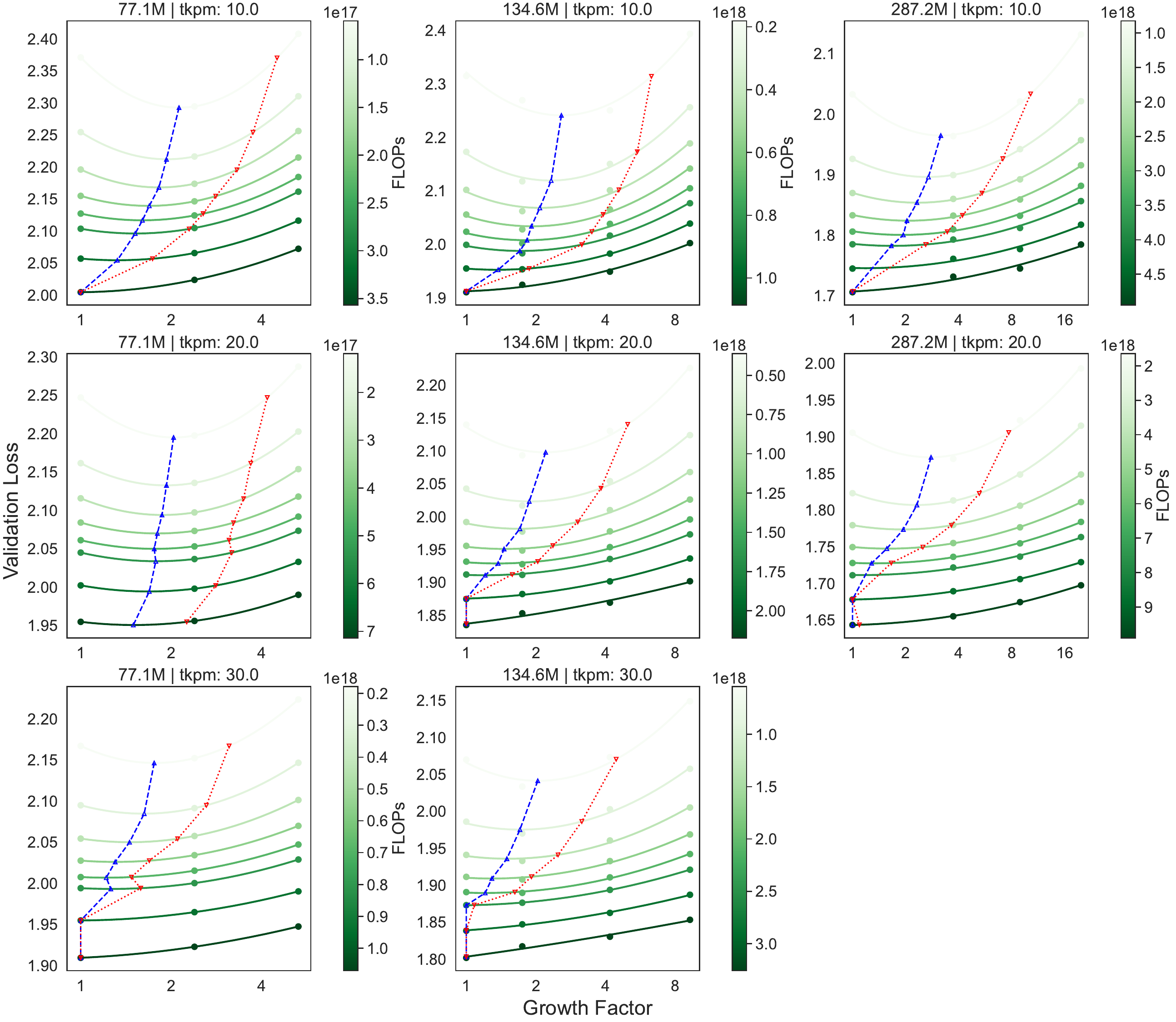}
    \caption{IsoFLOPs for \nton{} in the LLM setting. Blue and red lines denote $\gopt$ and $\gub$ respectively, as in \cref{fig:isoflops-main}.}
    \label{fig:isoflops_llm_n2n_app}
\end{figure}

\begin{figure}[htbp]
    \centering
    \includegraphics[width=\linewidth]{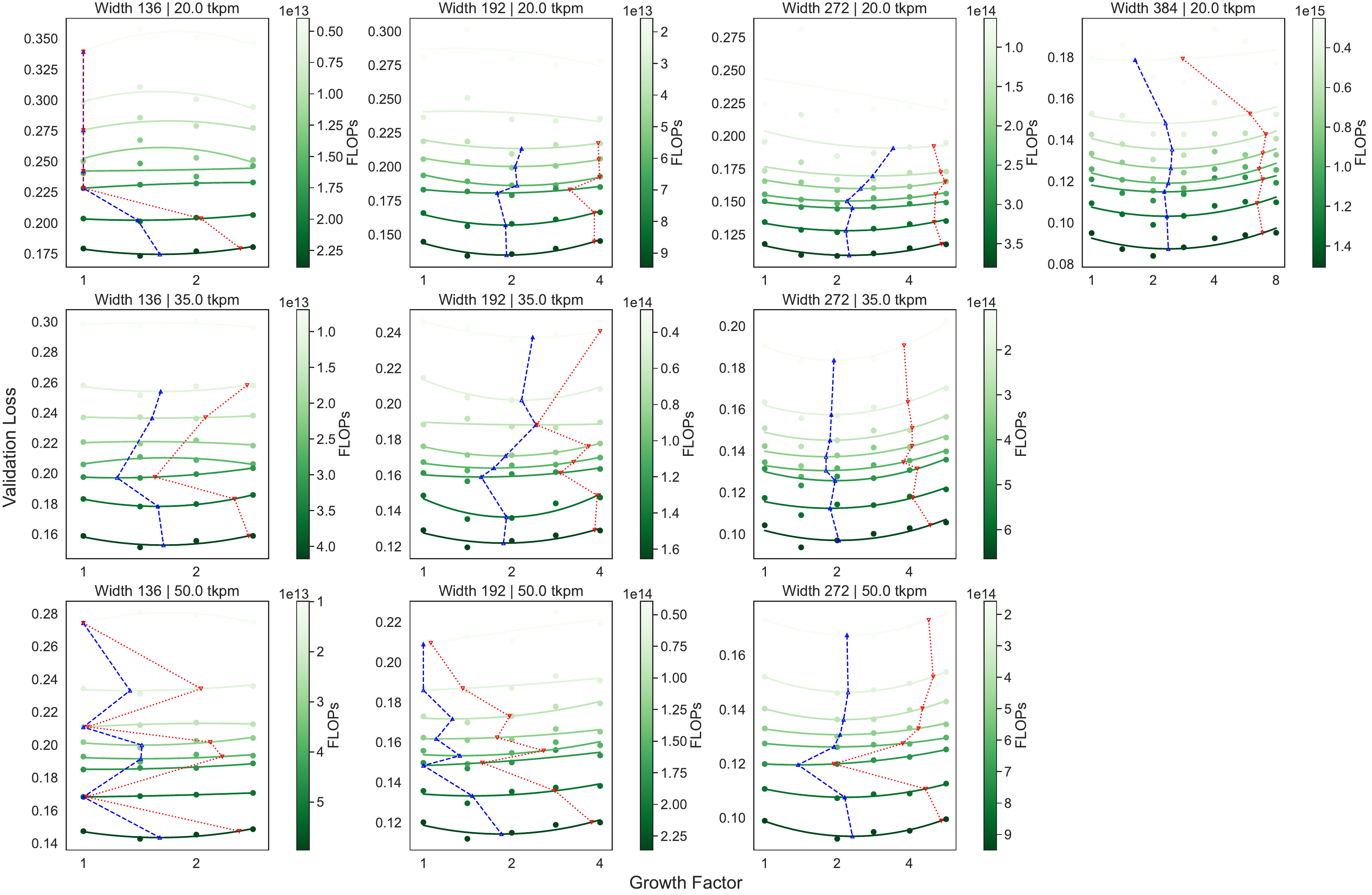}
    \caption{IsoFLOPs for \ours{} in the MLP setting. Blue and red lines denote $\gopt$ and $\gub$ respectively, as in \cref{fig:isoflops-main}.}
    \label{fig:isoflops_mlp_szp_app}
\end{figure}

\begin{figure}%
    \centering
    \includegraphics[width=\linewidth]{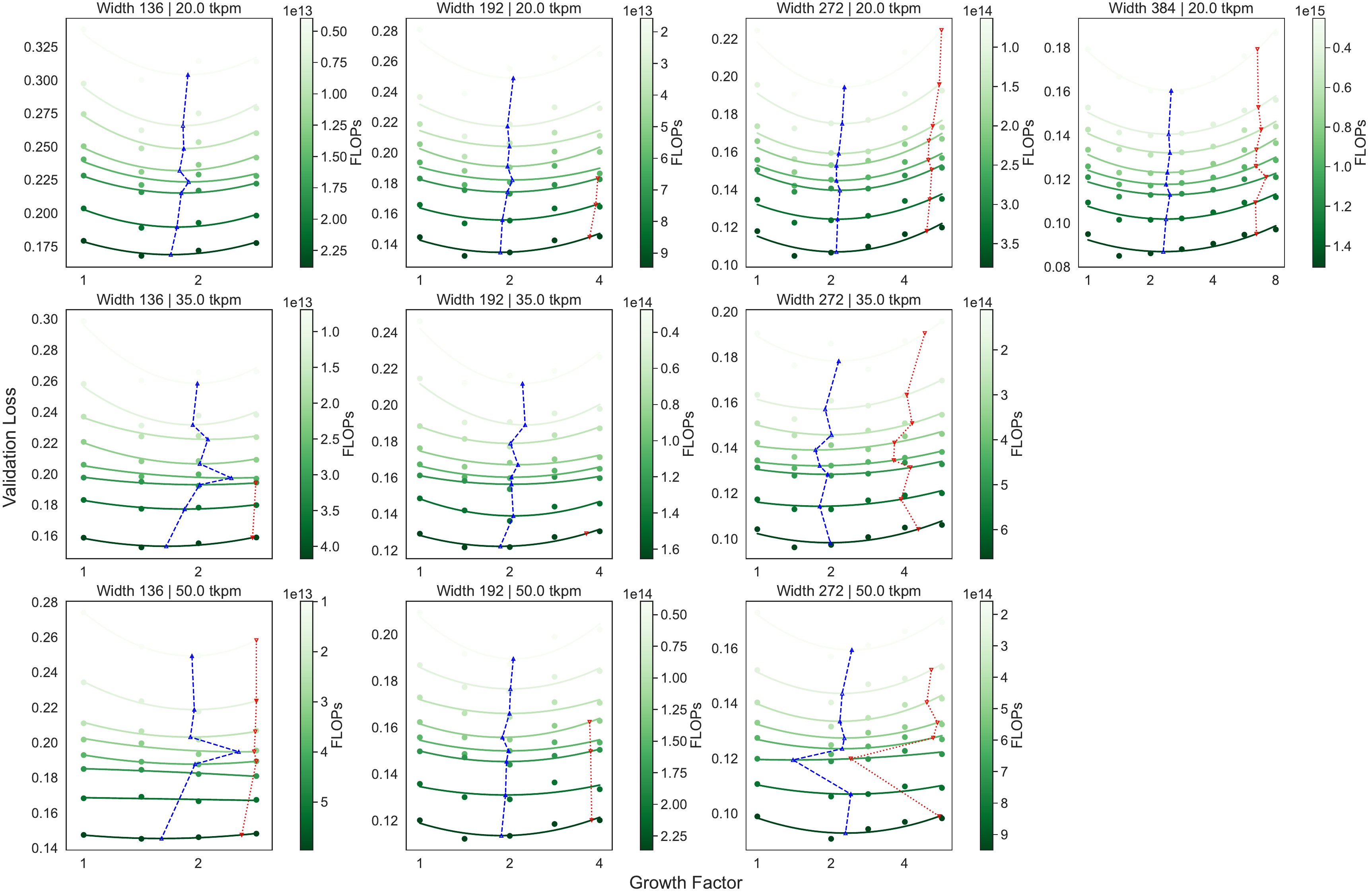}
    \caption{IsoFLOPs for \nton{} in the MLP setting. Blue and red lines denote $\gopt$ and $\gub$ respectively, as in \cref{fig:isoflops-main}.
    $\gub$ is omitted where it falls outside the range of collected data to avoid extrapolation.
    }
    \label{fig:isoflops_mlp_n2n_app}
\end{figure}

\paragraph{Extended Results.}
We extend the IsoFLOPs plots from \Cref{fig:isoflops-main} by including more target scales and token budgets.
\Cref{fig:isoflops_llm_szp_app,fig:isoflops_llm_n2n_app} show the results for \ours{} and \nton{} in the LLM setting, while \cref{fig:isoflops_mlp_szp_app,fig:isoflops_mlp_n2n_app} show the corresponding results for the MLP setting.
Similar to \Cref{fig:isoflops-main}, we can observe the existence of $\gub$ consistently across scales and budgets.
For \nton{} in the MLP setting, the combination of strong performance and the limited range of growth factors at smaller target scales means $\gub$ occasionally falls outside the observed data and is omitted.

\paragraph{Including the Base Compute.}
Our primary analysis assumes that the base model is given and therefore excludes its training compute. 
Prior work has shown that the compute invested in pretraining the base model influences the effectiveness of subsequent model reuse \citep{liew-arxiv25a}. 
We additionally investigate how accounting for the base-model training compute $C_\text{base}$ affects our IsoFLOP analysis, as reported in \Cref{tab:isoflop_basecompute}. 
Including $C_\text{base}$ shifts both $\gub$ and $\gopt$ toward smaller growth factors, with both reaching $g=1$ at the two largest budgets. 
Thus, while accounting for base-model compute changes the location of $\gub$ and can eliminate the advantage of warmstarting, our central observation remains unchanged: the range of growth factors for which warmstarting is beneficial is bounded.

\paragraph{Bootstrap Uncertainty.}
We use bootstrap resampling to quantify uncertainty in $\gopt$ and $\gub$. For each FLOPs budget, we resample the $g>1$ points with replacement and refit the quadratic in $\log g$, keeping Scratch ($g=1$) fixed as a common reference. 
\Cref{tab:bootstrap_llm} reports means and sample standard deviations from 1,000 valid bootstrap fits. 
We omit estimates where the full-data fit yields $\gopt>\gub$. Uncertainty varies across methods, target scales, and compute budgets, without establishing uniformly greater precision for $\gub$. 
While $\gopt$ specifies a growth factor to target, $\gub$ marks the upper end of a beneficial range. 
This flexibility is a practical benefit of reporting $\gub$, as practitioners can choose conservatively within this range without needing to match the boundary precisely.

\paragraph{Reliability of $\gopt$.} 
Note that following \citet{du-neurips24a}, $g=1$ corresponds to training from scratch rather than warmstarting from a same-sized model.
The quadratic fit assumes a minimum, i.e. $\gopt$, but this minimum may be an artifact of this conceptual mismatch: warmstarting from a same-sized model could perform comparably to or better than warmstarting at slightly larger growth factors, and we do not sample densely enough at low growth factors to distinguish these cases.
We therefore consider $\gub$ the more practically relevant quantity, while acknowledging that both estimates are subject to fitting uncertainty.

\paragraph{Toward a Unified Scaling Law.} Our IsoFLOP analysis models the effect of the growth factor $g$ on loss, while our scaling law fits capture the joint effect of model size $N$ and training budget $D$. 
Since both can be modeled independently and predictably, a unified parametric form $L(N,D,g)$ that captures all three axes in a single scaling law is a natural next step.
We consider this a promising direction for future work.

\begin{table}[t]
\caption{
Bootstrap estimates (mean $\pm$ sample standard deviation) of $\gopt$ and $\gub$ for SZP and Net2Net at 20 tokens/parameter, using 1,000 valid resampled quadratic IsoFLOP fits per budget with Scratch held fixed.
Dashes indicate omitted estimates where the full-data quadratic opens downward or yields $\gopt>\gub$.
}
\label{tab:bootstrap_llm}
\centering
\begin{tabular}{lcccc}
\toprule
& \multicolumn{2}{c}{\texttt{SZP}} & \multicolumn{2}{c}{\texttt{Net2Net}} \\
\cmidrule(lr){2-3} \cmidrule(lr){4-5}
FLOPs & $g_{\text{opt}}$ & $g_{\text{upper}}$ & $g_{\text{opt}}$ & $g_{\text{upper}}$ \\
\midrule
\multicolumn{5}{l}{LLM - Target 134.6M} \\
$3.6\times10^{17}$ & $2.61 \pm 0.19$ & $6.94 \pm 1.00$ & $2.20 \pm 0.08$ & $4.92 \pm 0.35$ \\
$6.2\times10^{17}$ & $2.46 \pm 0.15$ & $6.17 \pm 0.77$ & $1.94 \pm 0.11$ & $3.81 \pm 0.40$ \\
$8.8\times10^{17}$ & $2.38 \pm 0.13$ & $5.74 \pm 0.62$ & $1.74 \pm 0.10$ & $3.07 \pm 0.36$ \\
$1.1\times10^{18}$ & $2.30 \pm 0.11$ & $5.34 \pm 0.52$ & $1.53 \pm 0.11$ & $2.40 \pm 0.33$ \\
$1.4\times10^{18}$ & $2.26 \pm 0.09$ & $5.18 \pm 0.43$ & $1.43 \pm 0.11$ & $2.10 \pm 0.29$ \\
$1.7\times10^{18}$ & $2.20 \pm 0.08$ & $4.92 \pm 0.37$ & $1.26 \pm 0.11$ & $1.63 \pm 0.24$ \\
$1.9\times10^{18}$ & $2.17 \pm 0.06$ & $4.71 \pm 0.25$ & $1.00 \pm 0.00$ & $1.00 \pm 0.01$ \\
$2.2\times10^{18}$ & $2.11 \pm 0.05$ & $4.49 \pm 0.18$ & $1.00 \pm 0.00$ & $1.00 \pm 0.00$ \\
\midrule
\multicolumn{5}{l}{LLM - Target 287.2M} \\
$1.6\times10^{18}$ & $3.48 \pm 0.08$ & $12.04 \pm 0.53$ & $2.74 \pm 0.11$ & $7.55 \pm 0.56$ \\
$2.8\times10^{18}$ & $3.02 \pm 0.09$ & $9.05 \pm 0.50$ & $2.31 \pm 0.03$ & $5.28 \pm 0.14$ \\
$4.0\times10^{18}$ & $2.78 \pm 0.19$ & $7.58 \pm 1.07$ & $1.91 \pm 0.01$ & $3.67 \pm 0.04$ \\
$5.2\times10^{18}$ & $2.60 \pm 0.23$ & $6.62 \pm 1.25$ & $1.58 \pm 0.04$ & $2.50 \pm 0.13$ \\
$6.4\times10^{18}$ & $2.46 \pm 0.28$ & $5.95 \pm 1.42$ & $1.30 \pm 0.08$ & $1.66 \pm 0.21$ \\
$7.5\times10^{18}$ & $2.29 \pm 0.38$ & $5.21 \pm 1.74$ & -- & -- \\
$8.7\times10^{18}$ & $2.19 \pm 0.43$ & $4.80 \pm 1.88$ & $1.00 \pm 0.00$ & $1.01 \pm 0.01$ \\
$9.9\times10^{18}$ & $2.00 \pm 0.57$ & $4.18 \pm 2.21$ & $1.00 \pm 0.00$ & $1.02 \pm 0.02$ \\
\midrule
\multicolumn{5}{l}{MLP - Target width 272} \\
$6.3\times10^{13}$ & -- & -- & $2.24 \pm 0.16$ & $4.94 \pm 0.69$ \\
$1.1\times10^{14}$ & $3.98 \pm 1.48$ & $3.26 \pm 1.40$ & $2.22 \pm 0.17$ & $4.78 \pm 0.70$ \\
$1.5\times10^{14}$ & $3.50 \pm 1.36$ & $3.79 \pm 1.20$ & $2.15 \pm 0.16$ & $4.80 \pm 0.69$ \\
$2.0\times10^{14}$ & $2.69 \pm 0.71$ & $5.30 \pm 0.25$ & $2.11 \pm 0.14$ & $4.78 \pm 0.63$ \\
$2.4\times10^{14}$ & $2.24 \pm 0.44$ & $4.96 \pm 0.53$ & $2.10 \pm 0.13$ & $4.76 \pm 0.61$ \\
$2.9\times10^{14}$ & -- & -- & $2.17 \pm 0.11$ & $4.93 \pm 0.50$ \\
$3.3\times10^{14}$ & $2.16 \pm 0.10$ & $4.87 \pm 0.51$ & $2.11 \pm 0.14$ & $4.79 \pm 0.66$ \\
$3.8\times10^{14}$ & $2.22 \pm 0.13$ & $5.08 \pm 0.61$ & $2.09 \pm 0.13$ & $4.67 \pm 0.57$ \\
\midrule
\multicolumn{5}{l}{MLP - Target width 384} \\
$2.5\times10^{14}$ & $2.58 \pm 2.04$ & $3.16 \pm 1.99$ & $2.51 \pm 0.05$ & $6.42 \pm 0.33$ \\
$4.3\times10^{14}$ & $2.26 \pm 1.08$ & $5.06 \pm 1.81$ & $2.44 \pm 0.11$ & $6.38 \pm 0.57$ \\
$6.1\times10^{14}$ & $2.46 \pm 0.80$ & $6.11 \pm 1.52$ & $2.47 \pm 0.13$ & $6.53 \pm 0.75$ \\
$7.9\times10^{14}$ & $2.45 \pm 0.81$ & $6.32 \pm 1.39$ & $2.41 \pm 0.12$ & $6.27 \pm 0.63$ \\
$9.7\times10^{14}$ & $2.34 \pm 0.63$ & $6.08 \pm 1.33$ & $2.40 \pm 0.11$ & $6.21 \pm 0.59$ \\
$1.1\times10^{15}$ & $2.33 \pm 0.79$ & $5.99 \pm 1.42$ & $2.48 \pm 0.17$ & $6.86 \pm 0.95$ \\
$1.3\times10^{15}$ & $2.33 \pm 0.27$ & $6.16 \pm 1.05$ & $2.38 \pm 0.12$ & $6.18 \pm 0.64$ \\
$1.5\times10^{15}$ & $2.40 \pm 0.18$ & $6.57 \pm 1.04$ & $2.34 \pm 0.14$ & $6.21 \pm 0.73$ \\
\bottomrule
\end{tabular}
\end{table}

\begin{table}[t]
\caption{Effect of including the base-model training compute $C_{\text{base}}$ in the IsoFLOP analysis, for \texttt{SZP} on the 287M LLM at 20 \textit{tokens/parameter}. With $C_{\text{base}}$ included, $g_{\text{upper}}$ shifts to smaller growth factors and collapses to $1$ at the largest budgets.}
\label{tab:isoflop_basecompute}
\centering
\begin{tabular}{lcccc}
\toprule
& \multicolumn{2}{c}{$g_{\text{opt}}$} & \multicolumn{2}{c}{$g_{\text{upper}}$} \\
\cmidrule(lr){2-3} \cmidrule(lr){4-5}
FLOPs & with $C_{\text{base}}$ & without & with $C_{\text{base}}$ & without \\
\midrule
$1.65\times10^{18}$ & 3.03 & 3.52 & 8.27 & 12.39 \\
$2.83\times10^{18}$ & 2.46 & 2.99 & 5.12 & 8.78 \\
$4.01\times10^{18}$ & 2.35 & 2.77 & 4.41 & 7.12 \\
$5.18\times10^{18}$ & 2.28 & 2.61 & 4.15 & 6.13 \\
$6.36\times10^{18}$ & 2.25 & 2.53 & 4.03 & 5.69 \\
$7.54\times10^{18}$ & 2.25 & 2.46 & 3.97 & 5.28 \\
$8.72\times10^{18}$ & 1.00 & 2.42 & 1.00 & 5.04 \\
$9.90\times10^{18}$ & 1.00 & 2.28 & 1.00 & 4.61 \\
\bottomrule
\end{tabular}
\end{table}


\subsection{Scaling Law Analysis}
\label{app:scaling-fit}
\paragraph{Fitted Parameters.}
\Cref{tab:scaling-fit} reports the fitted parameters and fit quality for the scaling laws shown in \cref{fig:llm_N_D_ratio}.
For reference, we repeat the parametric form.
Following Approach 3 of~\citet{hoffmann-neurips22a}, we model the loss as a function of the number of parameters $N$ and training tokens $D$:
\begin{equation*}
L(N, D) = E + \frac{A}{N^{\alpha}} + \frac{B}{D^{\beta}},
\end{equation*}
where $E, A, \alpha, B, \beta$ are fit to the data. 
The resulting fits capture the data well, with all $R^2$ values exceeding $0.99$.

\begin{table}[ht]
\centering
\caption{
The fitted parameters and $R^2$ as a quality-of-fit metric for the scaling law equations of Figure~\ref{fig:llm_N_D_ratio}.
}
\label{tab:scaling-fit}
\begin{tabular}{lllllll}
\toprule
& \multicolumn{3}{l}{LLM} & \multicolumn{3}{l}{MLP}\\
 & \ours{} & Net2Net & Scratch & \ours{} & Net2Net & Scratch \\
\midrule
$A$ & $1038$ & $1739$ & $18807$ & $10484$ & $424$ & $1437$ \\
$\alpha$ & $0.439$ & $0.474$ & $0.629$ & $1.027$ & $0.737$ & $0.908$ \\
$B$ & $68$ & $194$ & $96$ & $774$ & $427$ & $87$ \\
$\beta$ & $0.257$ & $0.311$ & $0.267$ & $0.578$ & $0.555$ & $0.424$ \\
$E$ & $1.278$ & $1.377$ & $1.395$ & $0.075$ & $0.074$ & $0.065$ \\
\midrule
$R^2$ & $0.9999$ & $0.9929$ & $0.9922$ & $0.9994$ & $0.9929$ & $0.9922$ \\
\bottomrule
\end{tabular}
\end{table}

\begin{figure}[thbp]
    \centering
    \includegraphics[width=\linewidth]{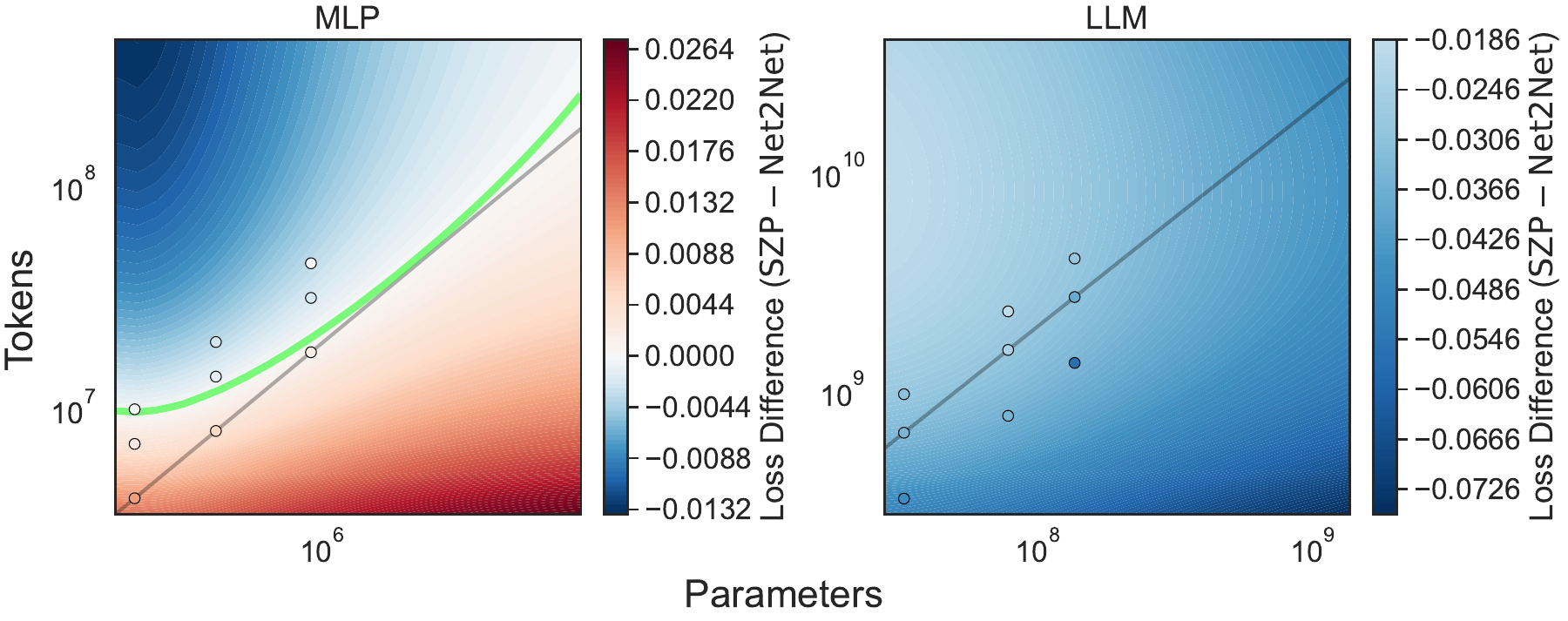}
    \caption{Isolines for \textit{loss difference} comparing \ours{} and \nton{} for growth factor $g\approx2$ ({\color{blue}blue regions} indicate \ours{} is better; {\color{red}red} indicates \nton{} is better; the {\color{green}green line} marks the boundary where both perform equally).
    In the LLM setting, \ours{} is predicted to outperform \nton{} across all configurations. In the MLP setting, \nton{} is competitive at lower token budgets, where its function-preservation properties are most beneficial, but \ours{} is predicted to perform better at larger token budgets.}
    \label{fig:scaling-law-comparison-warm}
\end{figure}

\paragraph{Further Applications.}
So far, we have used scaling-law fits to characterize when warmstarting is effective compared with training from scratch. 
The same approach can also be used for comparing \WS{} methods against each other.
In \Cref{fig:scaling-law-comparison-warm}, we compare \ours{} and \nton{} by plotting isolines of the difference between their fitted loss surfaces across model sizes and training budgets.
In the LLM setting, \ours{} is predicted to outperform \nton{} across all configurations. 
In the MLP setting, \nton{} is competitive at low token budgets, where its function-preservation properties are most beneficial, but \ours{} is predicted to perform better at larger token budgets. 
This aligns with the design of each method: \nton{} preserves the base model's function, providing an immediate advantage, while \ours{} enables the use of new neurons (see \cref{app:interpretability}), which becomes increasingly beneficial as the training budget grows and the model has more room to adapt.
More broadly, this demonstrates that scaling law fits enable a richer comparison between warmstarting methods than evaluations at a single training budget, capturing how each method's effectiveness varies with both model size and training duration.

\newpage

\section*{NeurIPS Paper Checklist}

\begin{enumerate}

\item {\bf Claims}
    \item[] Question: Do the main claims made in the abstract and introduction accurately reflect the paper's contributions and scope?
    \item[] Answer: \answerYes{} %
    \item[] Justification: The empirical results in~\Cref{sec:empirical} support the main claims, and~\Cref{sec:conclusion} summarizes the contributions and scope.

\item {\bf Limitations}
    \item[] Question: Does the paper discuss the limitations of the work performed by the authors?
    \item[] Answer: \answerYes{} %
    \item[] Justification: Limitations are discussed explicitly in the \Cref{sec:conclusion}.

\item {\bf Theory assumptions and proofs}
    \item[] Question: For each theoretical result, does the paper provide the full set of assumptions and a complete (and correct) proof?
    \item[] Answer: \answerNA{} %
    \item[] Justification: The work is primarily empirical. \Cref{sec:methodology} gives the method reasoning, and Appendix,~\Cref{app:szp-theory} provides theoretical intuition rather than formal theoretical results.

    \item {\bf Experimental result reproducibility}
    \item[] Question: Does the paper fully disclose all the information needed to reproduce the main experimental results of the paper to the extent that it affects the main claims and/or conclusions of the paper (regardless of whether the code and data are provided or not)?
    \item[] Answer: \answerYes{} %
    \item[] Justification: The codebase concerning the model training can be accessed via 
    \url{https://github.com/Neeratyoy/warmstarting_exps}
~\Cref{sec:empirical}, \Cref{app:synthetic-regression}, and \Cref{app:llm-details} provide complete details of all public resources and setups used.

\item {\bf Open access to data and code}
    \item[] Question: Does the paper provide open access to the data and code, with sufficient instructions to faithfully reproduce the main experimental results, as described in supplemental material?
    \item[] Answer: \answerYes{}{} %
    \item[] Justification: Detailed in ~\Cref{sec:empirical}, \Cref{app:synthetic-regression}, and \Cref{app:llm-details}.

\item {\bf Experimental setting/details}
    \item[] Question: Does the paper specify all the training and test details (e.g., data splits, hyperparameters, how they were chosen, type of optimizer) necessary to understand the results?
    \item[] Answer: \answerYes{} %
    \item[] Justification: Training and evaluation details are provided in \Cref{sec:empirical}; \Cref{app:synthetic-regression}, and \Cref{app:llm-details} of the Appendix.

\item {\bf Experiment statistical significance}
    \item[] Question: Does the paper report error bars suitably and correctly defined or other appropriate information about the statistical significance of the experiments?
    \item[] Answer: \answerNo{} %
    \item[] Justification: Due to the compute requirements, experiments are not repeated with error bars. The study uses controlled settings and compares relative performance.

\item {\bf Experiments compute resources}
    \item[] Question: For each experiment, does the paper provide sufficient information on the computer resources (type of compute workers, memory, time of execution) needed to reproduce the experiments?
    \item[] Answer: \answerYes{} %
    \item[] Justification: The Appendix,~\Cref{{app:compute}} reports the compute resources, and the figures report the FLOPs for the corresponding runs.
    
\item {\bf Code of ethics}
    \item[] Question: Does the research conducted in the paper conform, in every respect, with the NeurIPS Code of Ethics \url{https://neurips.cc/public/EthicsGuidelines}?
    \item[] Answer: \answerYes{} %
    \item[] Justification: The work uses public/open-source resources and studies LLM pretraining efficiency. To the best of our knowledge, it does not violate the NeurIPS Code of Ethics.

\item {\bf Broader impacts}
    \item[] Question: Does the paper discuss both potential positive societal impacts and negative societal impacts of the work performed?
    \item[] Answer: \answerNA{} %
    \item[] Justification: The work is methodological and targets LLM pretraining efficiency, with no direct societal impact beyond possible indirect compute savings.
    
\item {\bf Safeguards}
    \item[] Question: Does the paper describe safeguards that have been put in place for responsible release of data or models that have a high risk for misuse (e.g., pre-trained language models, image generators, or scraped datasets)?
    \item[] Answer: \answerNA{} %
    \item[] Justification: The work proposes a training technique and does not release models or datasets intended for direct use.

\item {\bf Licenses for existing assets}
    \item[] Question: Are the creators or original owners of assets (e.g., code, data, models), used in the paper, properly credited and are the license and terms of use explicitly mentioned and properly respected?
    \item[] Answer: \answerYes{} %
    \item[] Justification: Suitable license selected during submission.

\item {\bf New assets}
    \item[] Question: Are new assets introduced in the paper well documented and is the documentation provided alongside the assets?
    \item[] Answer: \answerNA{} %
    \item[] Justification: The paper does not introduce or release new assets.

\item {\bf Crowdsourcing and research with human subjects}
    \item[] Question: For crowdsourcing experiments and research with human subjects, does the paper include the full text of instructions given to participants and screenshots, if applicable, as well as details about compensation (if any)? 
    \item[] Answer: \answerNA{} %
    \item[] Justification: The paper does not involve crowdsourcing experiments or research with human subjects.

\item {\bf Institutional review board (IRB) approvals or equivalent for research with human subjects}
    \item[] Question: Does the paper describe potential risks incurred by study participants, whether such risks were disclosed to the subjects, and whether Institutional Review Board (IRB) approvals (or an equivalent approval/review based on the requirements of your country or institution) were obtained?
    \item[] Answer:  \answerNA{} %
    \item[] Justification: The paper does not involve study participants or human-subject research.

\item {\bf Declaration of LLM usage}
    \item[] Question: Does the paper describe the usage of LLMs if it is an important, original, or non-standard component of the core methods in this research? Note that if the LLM is used only for writing, editing, or formatting purposes and does \emph{not} impact the core methodology, scientific rigor, or originality of the research, declaration is not required.
    \item[] Answer: \answerNA{} %
    \item[] Justification: Used primarily for grammar checks, improving visualization aesthetics, and code assistance.

\end{enumerate}

\end{document}